\documentclass[10pt,twocolumn,letterpaper]{article}

\usepackage{cvpr}              % To produce the CAMERA-READY version
\usepackage[utf8]{inputenc}
\usepackage[accsupp]{axessibility} 
\usepackage{fontenc}
\usepackage{amsmath}
\usepackage{amssymb}
\usepackage{graphicx}
\usepackage{enumitem}
\usepackage[margin=1in]{geometry}
\usepackage{booktabs}
\usepackage{lmodern}
\usepackage{parskip}
\usepackage{times}
\usepackage{epsfig}
\usepackage{graphicx}
\usepackage{longtable}
\definecolor{cvprblue}{rgb}{0.21,0.49,0.74}
\usepackage[pagebackref,breaklinks,colorlinks,allcolors=cvprblue]{hyperref}

\title{ProGAL-VLA: Grounded Alignment through Prospective Reasoning in Vision-Language-Action Models}

\author{Nastaran Darabi, and Amit Ranjan Trivedi\\
University of Illinois Chicago, IL, USA\\
{\tt\small (ndarab2, amitrt)@uic.edu}\\
 Project page: \hyperlink{https://nstrndrbi.github.io/ProGAL}{https://nstrndrbi.github.io/ProGAL}
}

\begin{document}
\maketitle

\setlength{\textfloatsep}{10pt plus 1pt minus 1pt}
\setlength{\floatsep}{8pt plus 1pt minus 1pt}
\setlength{\intextsep}{10pt plus 1pt minus 1pt}
\setlength{\parskip}{3pt}

\begin{abstract}
Vision language action (VLA) models enable generalist robotic agents but often exhibit language ignorance, relying on visual shortcuts and remaining insensitive to instruction changes. We present Prospective Grounding and Alignment VLA (ProGAL-VLA), which constructs a 3D entity-centric graph (GSM), uses a slow planner to produce symbolic sub-goals, and aligns them with grounded entities via a Grounding Alignment Contrastive (GAC) loss. All actions are conditioned on a verified goal embedding $g_t$, whose attention entropy provides an intrinsic ambiguity signal. On LIBERO-Plus, ProGAL-VLA increases robustness under robot perturbations from 30.3 to 71.5 percent, reduces language ignorance by 3x-4x, and improves entity retrieval from 0.41 to 0.71 Recall@1. On the Custom Ambiguity Benchmark, it reaches AUROC 0.81 (vs.\ 0.52), AUPR 0.79, and raises clarification on ambiguous inputs from 0.09 to 0.81 without harming unambiguous success. The verification bottleneck increases mutual information of language-actions, the GAC loss imposes an entity-level InfoNCE bound, and attention entropy yields calibrated selective prediction, indicating that explicit verified grounding is an effective path toward instruction-sensitive, ambiguity-aware agents.
\end{abstract}

\section{Introduction}

Vision language action (VLA) models have advanced generalist robotic agents capable of open vocabulary reasoning and manipulation~\cite{kim2024openvla,ma2024survey}. Transformer based policies such as RT 2~\cite{zitkovich2023rt}, PaLM E~\cite{driess2023palm}, Gato~\cite{reed2022generalist}, and open source systems including OpenVLA~\cite{kim2024openvla}, Eureka~\cite{balachandran2024eureka}, and $\pi_{0.5}$~\cite{intelligence2025pi_} unify perception, language, and control through multimodal pretraining~\cite{li2025_250815201}. However, evaluations such as LIBERO-Plus~\cite{fei2025libero} reveal two persistent failures: (i) \textit{language ignorance}, where policies rely on visual priors instead of instruction semantics, and (ii) \textit{robotic instability}, where control degrades once semantic reasoning perturbs visuomotor coordination. 

Models that perform well on linguistic reasoning often underperform in the robot category, thus exposing a structural mismatch between symbolic intent and embodied control~\cite{toyoda2021_210408521}. This mismatch arises because most VLAs fuse modalities through shallow concatenation, allowing the language embedding to modulate vision without ensuring that the inferred \emph{symbolic goal} corresponds to an \emph{actionable entity} in the 3D scene. As a result, control layers operate on unverified representations, producing semantic and physical failures~\cite{zhu2024_241003659, lu2023_230904041, han2025_250402477}.

\begin{figure}[t]
    \centering
    \includegraphics[width=\linewidth]{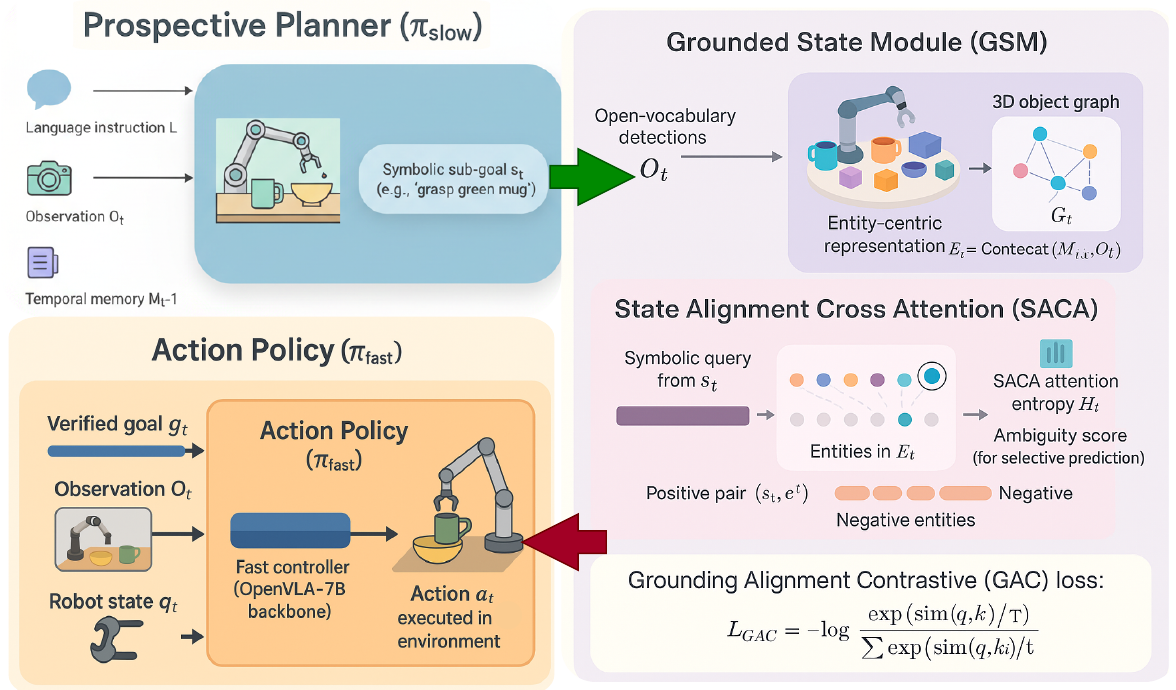}
\caption{\textbf{Overview of ProGAL-VLA}. Language instruction $L$ and observation $O_t$ are processed by the Prospective Planner and the Grounded State Module (GSM). The State Alignment Cross Attention (SACA) module verifies alignment between the symbolic sub-goal $s_t$ and 3D entities $E_t$, producing a verified goal embedding $g_t$ for the Action Policy ($\pi_{\text{fast}}$). The Grounding Alignment Contrastive (GAC) objective enforces correct symbolic to entity binding during training.}
    \label{fig:overview}
\end{figure}

We address this with ProGAL-VLA (Prospective Grounding and Alignment VLA), a hierarchical architecture that imposes explicit grounding verification before any action is executed (Fig.~\ref{fig:overview}). The core idea is a verification bottleneck that ensures linguistic intent is both semantically consistent and physically actionable within the current 3D scene. ProGAL-VLA separates reasoning and control into two asynchronous components: (i) the \textit{Prospective Planner} ($\pi_{\text{slow}}$), which predicts a symbolic sub-goal $s_t$ from the instruction and context, and (ii) the \textit{Grounded State Module} (GSM), which constructs an entity-centric 3D representation of the environment. 

These components interact through a \textit{State Alignment Cross Attention} (SACA) mechanism that binds the symbolic goal to the grounded entities, producing a verified goal embedding $g_t$. The downstream Action Policy ($\pi_{\text{fast}}$) conditions only on $g_t$, ensuring that each action is supported by perceptual evidence. Training uses a \textit{Grounding Alignment Contrastive} (GAC) objective that aligns symbolic sub-goals with grounded 3D entities and penalizes queries that cannot be verified. This structure improves both semantic grounding and robotic robustness: the verified embedding $g_t$ stabilizes low level control and preserves consistent actuation under visual perturbations such as camera shifts, layout changes, and lighting variation.

Empirical results on LIBERO-Plus~\cite{fei2025libero} and our Custom Ambiguity Benchmark (CAB) show that ProGAL-VLA mitigates language indifference while substantially improving embodied robustness, with the largest gains in the robot and camera categories. On LIBERO-Plus, robustness under robot perturbations increases from 30.3 to 71.5 percent, language indifference drops by 3x-4x across simple, spatial, and relational instructions, and entity retrieval improves from 0.41 to 0.71 Recall@1 (N=8). On CAB, ProGAL-VLA attains AUROC 0.81 (vs. 0.52), AUPR 0.79, and raises clarification on ambiguous inputs from 0.09 to 0.81 without reducing unambiguous success.

\section{Relation to Prior Works}
\label{sec:related_work}

\noindent\textbf{Vision Language Action Models.}
Transformer-based VLA policies have enabled large-scale robotic manipulation. Gato~\cite{reed2022generalist} showed that a single model can span diverse tasks, and RT~1~\cite{brohan2022rt} demonstrated scalable sensorimotor control. RT~2~\cite{zitkovich2023rt} and PaLM~E~\cite{driess2023palm} transferred internet-scale semantics into robotic policies, enabling zero-shot generalization. These models remain monolithic and provide no mechanism to ensure that linguistic intent corresponds to perceptual evidence. Language acts as a weak conditioning signal, making behavior unstable when visual and textual cues conflict. ProGAL-VLA builds on such backbones through $\pi_{\text{fast}}$ but introduces explicit grounding and verification.

\noindent\textbf{Language Grounding in Robotics.}
Grounding failures grow with model scale. Fei et al.~\cite{fei2025libero} showed that many LIBERO-Plus policies are insensitive to instruction changes and rely on visual heuristics. Similar failures appear in long-horizon and compositional tasks~\cite{shi2025memoryvla, zhang2025pure, huang2025graphcot}. These findings motivate architectures that force linkage between symbolic intent and concrete scene structure. ProGAL-VLA introduces such a mechanism by requiring that the planner’s sub-goal be bound to a grounded entity before execution.

\noindent\textbf{Hierarchical and Dual Stream Architectures.}
Recent work separates high-level reasoning from low-level control using LLM planners~\cite{puertamerino2025_250108068, han2024_240414285}. SayCan~\cite{brohan2023can} uses an LLM to suggest sub-goals validated against a skill library, and Code as Policies~\cite{liang2022code} expresses plans as executable programs. These approaches improve long-horizon behavior but inject plans as text or features without verifying correspondence to the perceptual scene. ProGAL-VLA differs by introducing State Alignment Cross Attention, which serves as a verification bottleneck forcing the symbolic sub-goal from $\pi_{\text{slow}}$ to bind to a specific 3D entity before $\pi_{\text{fast}}$ can act.

\noindent\textbf{3D Scene Representation for Grounding.}
Structured, entity-centric representations improve grounding over 2D patch features that mix identity with pose and appearance. Prior work explores scene graphs~\cite{chang2021comprehensive}, slot based models such as Loc NERF~\cite{maggio2022loc}, and NeRF based perception~\cite{li2024gp}. GSM follows this direction by constructing a 3D object graph $G_t$ and a temporal memory $M_t$ to form entity disentangled representations $E_t$. This supports symbolic to entity binding for manipulation~\cite{mavridis2006grounded}.

\noindent\textbf{Contrastive Learning for Cross Modal Alignment.}
Contrastive learning aligns heterogeneous modalities, as in CLIP~\cite{radford2021learning} and robotics extensions such as R3M~\cite{nair2022r3m} and video-based models~\cite{radosavovic2023real, qian20243d}. Prior grounding work typically aligns text spans to 2D regions. ProGAL-VLA applies contrastive alignment to a finer level: GAC aligns symbolic sub-goal tokens with 3D entity embeddings, addressing the binding problem that limits accurate manipulation.

% \noindent\textbf{Ambiguity and Uncertainty in Robotic Vision.}
% Ambiguous language can lead to arbitrary actions unless the system can detect that grounding is underdetermined~\cite{yuan2025scene, qu2024chatvtg}. ProGAL-VLA uses SACA attention entropy as an intrinsic ambiguity signal. High entropy indicates that no single entity satisfies the symbolic query, providing a simple and calibrated mechanism for selective prediction and clarification decisions.

\section{Prospective Grounding for Vision Language Action Policies}

Existing VLA systems typically inject language into visuomotor pipelines through feature concatenation or shallow conditioning, encouraging ``lazy'' policies that rely on visual shortcuts instead of instruction semantics. ProGAL-VLA replaces this paradigm with a prospective grounding mechanism: no action is executed unless the inferred symbolic goal is verified to correspond to a specific perceptual entity. The architecture consists of four components: the Prospective Planner ($\pi_{\text{slow}}$), the Grounded State Module (GSM), the State Alignment Cross Attention (SACA), and the Action Policy ($\pi_{\text{fast}}$). The first two generate symbolic intent and 3D structure; SACA verifies their correspondence; and $\pi_{\text{fast}}$ acts only on verified goals.

\vspace{3pt}
\noindent\textbf{Prospective Planner ($\pi_{\text{slow}}$).}
The Prospective Planner predicts a symbolic sub-goal that represents the intended next step of the task:
\begin{equation}
s_t = \pi_{\text{slow}}(L, O_t, M_{t-1}),
\end{equation}
where $L$ is the instruction, $O_t$ the current observation, and $M_{t-1}$ a temporal memory. The planner focuses on semantic reasoning rather than fine-grained visuomotor details. Its output $s_t$ expresses what should be done next (e.g., ``grasp red block''), leaving the question of where this entity is located to the grounding stage. Notably, $\pi_{\text{slow}}$ performs no low-level reasoning or multi-step lookahead. It outputs a short symbolic template that normalizes linguistic variability (e.g. ``pick up the green mug'' to \textit{grasp\_green\_mug}), reducing lexical entropy and supplying a structured query for SACA rather than an additional control policy. Alignment is still driven primarily by GSM and SACA.

\vspace{3pt}
\noindent\textbf{Grounded State Module (GSM).} The GSM converts raw observations into an entity-centric 3D representation:
\begin{align}
G_t &= \{ e_1, e_2, \dots, e_n \}, \\
M_t &= \text{Update}(M_{t-1}, G_t), \\
E_t &= \text{Concat}(G_t, \text{Retrieve}(M_{t-1}, O_t)).
\end{align}
Each entity node $e_i$ encodes object-level visual features, geometric pose, and semantic descriptors. The memory $M_t$ maintains temporal continuity, enabling reasoning about occluded or partially observed objects. Unlike 2D patch features, this structured representation disentangles object identity from pose and appearance, providing the precision needed for symbolic binding. Notably, GSM is intentionally minimal: it lifts detector outputs to 3D, tracks a small set of entities, and maintains short-term temporal consistency. It does 
not perform semantic reasoning or long-horizon prediction; those are delegated to $\pi_{\text{slow}}$. GSM’s role is to expose the object-centric structure for 
SACA, not to replace a full perception stack.

\vspace{3pt}
\noindent\textbf{State Alignment Cross Attention (SACA).}
SACA performs the core grounding verification. The symbolic goal is embedded as a query
\(
Q = \text{Embed}_{\text{sym}}(s_t),
\)
and grounded entities are embedded as keys and values:
\begin{align}
K,V &= \text{Embed}_{\text{gnd}}(E_t), \\
g_t &= \text{Softmax}\!\left(\frac{QK^\top}{\sqrt{d}}\right)V.
\end{align}
The attention distribution links $s_t$ to specific 3D entities. Its entropy provides an intrinsic ambiguity signal: low entropy indicates a unique grounding, while high entropy reveals that the instruction cannot be resolved to a single entity. Only the verified goal embedding $g_t$ is passed to the controller. This bottleneck blocks actions that are semantically underdetermined or contradict the scene structure.

\vspace{3pt}
\noindent\textbf{Action Policy ($\pi_{\text{fast}}$).}
The policy receives the verified goal $g_t$, observation $O_t$, and robot state $q_t$:
\begin{equation}
a_t \sim \pi_{\text{fast}}(g_t, O_t, q_t),
\end{equation}
trained by imitation:
\begin{equation}
\mathcal{L}_{\text{action}} = \|a_t - a_t^*\|^2.
\end{equation}
Excluding raw language features from $\pi_{\text{fast}}$ ensures that actions depend solely on verified semantics and perceptual evidence rather than ungrounded linguistic cues.

\vspace{3pt}
\noindent\textbf{Grounding Alignment Contrastive (GAC) Loss.}
To enforce consistent sub-goal to entity correspondence, each symbolic–entity pair $(s_t, e^+)$ is used to form a contrastive objective. Let
\begin{equation}
q = \text{Embed}_{\text{sym}}(s_t), \qquad k^+ = \text{Embed}_{\text{gnd}}(e^+).
\end{equation}
The GAC loss is
\begin{equation}
\mathcal{L}_{\text{GAC}} = -\log
\frac{\exp(\text{sim}(q,k^+)/\tau)}
{\sum_i \exp(\text{sim}(q,k_i)/\tau)}.
\end{equation}
The overall training loss is
\begin{equation}
\mathcal{L}_{\text{total}} = \mathcal{L}_{\text{action}} + \lambda \mathcal{L}_{\text{GAC}}.
\end{equation}
GAC encourages the planner to generate sub-goals that correspond to actual entities and forces GSM to produce entity embeddings that remain discriminable under viewpoint and layout variation.

\vspace{3pt}
\noindent\textbf{Offline Generation of Alignment Pairs.}
We construct alignment pairs $(s_t, e^+)$ from demonstrations in three steps. 
\textit{(i) Symbolic segmentation:} A teacher VLM segments each demonstration into sub-goals, producing the symbolic plan $s_t$. 
\textit{(ii) Entity tracking:} The GSM builds 3D tracklets $\{T_{obj_i}\}$ for all objects in the scene. 
\textit{(iii) Spatio-temporal matching:} The grounded entity is selected by nearest-neighbor consistency at the end of the sub-goal:
\begin{equation}
    e^+ = \arg\min_i \|\text{pos}(T_{obj_i}, t_{\text{end}}) - \text{pos}(\text{gripper}, t_{\text{end}})\|.
\end{equation}
This procedure imposes a verified grounding bottleneck: every symbolic sub-goal must correspond to a physically reachable entity before the policy can act. This suppresses language-ignorance failure modes, avoids visually driven shortcuts, and stabilizes behavior under ambiguity. The decomposition into reasoning, grounding, and control is explicit and only connected through the verification step, yielding more instruction-sensitive VLA behavior.

Additionally, supervision remains weak. Entity identities and sub-goal labels are never annotated, and both the VLM segmentation and the matching process introduce noise and occasional inconsistencies. Thus the GAC loss does not rely on privileged information unavailable to baselines. Robust gains despite this noise (Table~\ref{core}) indicate that improvements stem from the verified grounding mechanism itself, not from cleaner supervision.

\section{Theoretical Foundations of Verified Grounding in VLA Models}
\label{sec:theory}

We analyze how the verification bottleneck, the GAC objective, and SACA entropy contribute to instruction sensitivity, entity-level grounding, and robustness under observation perturbations. The goal is not to provide asymptotic guarantees but to formalize why our architectural choices improve language grounding and stability in practice.

\vspace{3pt}
\noindent\textbf{Language influence via conditional mutual information.}
Let $(L,O_t,q_t,a_t)$ denote instruction, observation, robot state, and action under a fixed policy $\pi$. The extent to which actions depend on language beyond what is captured by $(O_t,q_t)$ is quantified by
\begin{equation}
\mathcal{I}_t^{\pi} \triangleq I(L; a_t \mid O_t, q_t),
\end{equation}
with $\mathcal{I}_t^{\pi}=0$ indicating language ignorance behavior. To compare across instruction spaces, we define
\begin{equation}
\Lambda^{\pi} \triangleq 1 - \frac{\mathbb{E}[\mathcal{I}_t^{\pi}]}{\log|\mathcal{L}|},
\label{eq:language-ignorance-index}
\end{equation}
where $\Lambda^{\pi}\approx 1$ implies minimal influence of language. The empirical language ignorance scores in Sec.~\ref{sec:results} estimate $\Lambda^{\pi}$ for inputs $(L,s_t,g_t)$.

\vspace{3pt}
\noindent\textbf{Effect of the verification bottleneck.}
ProGAL-VLA factors reasoning and control through

\begin{align}
s_t &= \pi_{slow}(L,O_t,M_{t-1}) \\
g_t &= \mathrm{SACA}(s_t,E_t) \\
a_t &\sim \pi_{fast}(g_t,O_t,q_t)
\end{align}

where $g_t$ is the verified embedding. The key structural assumption is:

\noindent\textbf{\textit{Assumption 1 (Verification bottleneck)}.}
\begin{equation}
a_t \perp (L,O_t,M_{t-1}) \mid (g_t,q_t).
\end{equation}

\noindent\textbf{\textit{Proposition 1.}}
Under Assumption 1,
\begin{equation}
I(L;a_t\mid O_t,q_t)
=
I(L;g_t\mid O_t,q_t)
-
I(L;g_t\mid a_t,O_t,q_t).
\label{eq:language-influence-decomposition}
\end{equation}

\noindent\emph{Interpretation.} Eq.~\eqref{eq:language-influence-decomposition} shows that the policy is sensitive to language whenever (i) the verified embedding $g_t$ varies meaningfully with $L$ given $(O_t,q_t)$ and (ii) $\pi_{\text{fast}}$ responds to this variation. Empirically, $g_t$ achieves the highest language influence among $(L,s_t,g_t)$, matching the sharp drop in language ignorance in Sec.~\ref{sec:results}.

\vspace{3pt}
\noindent\textbf{Entity-Level Information Bound from GAC.}
Let $s_t$ denote symbolic sub-goals and $e^+$ the corresponding grounded entity. With embeddings $q=\mathrm{Embed}_{\text{sym}}(s_t)$ and $k_i=\mathrm{Embed}_{\text{gnd}}(e_i)$, the GAC loss is
\begin{equation}
\mathcal{L}_{\text{GAC}}
=
-\log
\frac{\exp(\mathrm{sim}(q,k^+)/\tau)}
{\sum_{i=1}^{N}\exp(\mathrm{sim}(q,k_i)/\tau)}.
\label{eq:gac-tight}
\end{equation}

Let $S$ and $E$ denote random variables over symbolic sub-goals and grounded entities, and assume negatives are drawn from $p(E)$.

\noindent\textbf{\textit{Theorem 1 (InfoNCE-style lower bound)}.}
\begin{equation}
I(S;E)
\;\ge\;
\log N - \mathbb{E}[\mathcal{L}_{\text{GAC}}].
\label{eq:info-bound-gac}
\end{equation}

\noindent
\emph{Interpretation.} Minimizing $\mathcal{L}_{\text{GAC}}$ increases a lower bound on the mutual information between symbolic tokens and 3D entities. This explains the large gains in Recall@1 retrieval in Sec.~\ref{sec:results} across increasing candidate set sizes.

\vspace{3pt}
\noindent\textbf{Robustness to Visual Perturbations.}
Let $O_t' = T(O_t)$ for a transformation $T$ in a perturbation group $\mathcal{G}$. The grounding pipeline produces $g_t=\Psi(L,O_t,M_{t-1})$ and $g_t'=\Psi(L,O_t',M_{t-1})$.

\noindent\textbf{\textit{Assumption 2 (Lipschitz grounding).}}
\begin{equation}
    \|g_t' - g_t\|_2 \le L_{\Psi}\, d_{\mathcal{G}}(T,\mathrm{id}).
\end{equation}

\noindent\textbf{\textit{Assumption 3 (Lipschitz policy).}}
\begin{align}
\mathrm{TV}\bigl(\pi_{\text{fast}}(\cdot\mid g,O_t,q_t),
\pi_{\text{fast}}(\cdot\mid g',O_t,q_t)\bigr)\\
\le L_{\pi}\|g-g'\|_2.
\end{align}

\noindent\textbf{\textit{Proposition 2 (Action robustness)}.}
Under Assumptions 2--3,
\begin{align}
\mathrm{TV}\bigl(
\pi_{\text{fast}}(\cdot\mid g_t,O_t,q_t),
\pi_{\text{fast}}(\cdot\mid g_t',O_t',q_t)
\bigr)\\
\le
L_{\pi} L_{\Psi}\, d_{\mathcal{G}}(T,\mathrm{id}).
\label{eq:robustness-tight}
\end{align}

\noindent
\emph{Interpretation.} If $g_t$ changes smoothly under camera, layout, and lighting perturbations, the induced action distribution is provably stable. This aligns with the empirical robustness gains in Table~\ref{tab:libero-robustness}.

\noindent\textbf{Selective Prediction via SACA Entropy.}
SACA produces attention weights $\alpha_{t,i}$ over entities. The entropy
\begin{equation}
  H_t
  \;\triangleq\;
  - \sum_{i=1}^{n} \alpha_{t,i} \log \alpha_{t,i},
  \label{eq:saca-entropy}
\end{equation}

captures grounding ambiguity. Thresholding $H_t$ yields an abstaining policy with a Risk Coverage curve $\mathcal{C}$ that is strictly improved for ProGAL-VLA (Fig.~\ref{fig:risk_coverage}). This supports the calibrated ambiguity detection reported on CAB.

\medskip
\noindent\textbf{\textit{Remark.}} Additional analysis of the grounding margin over GSM nodes and its stability under attention perturbations is provided in the Appendix.

\section{Evaluation}
\label{sec:evaluation}
We evaluate ProGAL-VLA under a protocol designed to test three hypotheses: (1) the architecture reduces the language-ignorance failure observed in prior work; (2) integrating symbolic reasoning with entity-level grounding improves performance on long-horizon manipulation; and (3) the verification bottleneck produces emergent ambiguity detection and calibrated uncertainty.

\subsection{Experimental Setup}

\noindent\textbf{Base model.} The fast control policy $\pi_{\text{fast}}$ is instantiated with OpenVLA-7B, identical to all baselines. The slow planner $\pi_{\text{slow}}$ uses Qwen-2.5-VL-Instruct-7B, but it operates asynchronously and is invoked only \emph{once per episode} to produce a symbolic sub-goal template rather than at every control step. Thus, the per-step inference stack remains the same 7B OpenVLA backbone used by the baselines. This design isolates the contribution of the verified grounding pipeline, SACA, GSM, and the GAC objective, rather than confounding improvements with increased per-step model capacity.

\noindent\textbf{Perception.} Open-vocabulary detections and instance proposals are obtained from YOLO-World and lifted into the entity-centric 3D representation constructed by the GSM. The detector is used exactly as released, with no architectural or training modifications, ensuring that perception capacity is identical across ProGAL-VLA and all baselines.

\noindent\textbf{Fine-tuning.} All components start from publicly released pretrained weights and are fine-tuned end-to-end on LIBERO-Plus~\cite{fei2025libero} and our Custom Ambiguity Benchmark (CAB). Ambiguity supervision is constructed via the post-hoc spatio-temporal alignment pipeline (Sec.~3.5), which yields the $(s_t, e^+)$ pairs needed to optimize both the action loss $\mathcal{L}_{\text{action}}$ and the contrastive grounding objective $\mathcal{L}_{\text{GAC}}$.

\subsection{Evaluation Benchmarks}

We evaluate on both established and newly introduced benchmarks to test reasoning, grounding fidelity, robustness, and ambiguity awareness.

\textbf{LIBERO-Plus.}
LIBERO-Plus~\cite{fei2025libero} is the benchmark on which the language ignorance phenomenon was first shown. It therefore serves as the primary testbed for assessing whether ProGAL-VLA, in particular SACA and the mutual information-based $\mathcal{L}_{\text{GAC}}$ (Sec.~\ref{sec:theory}), reduces invariance to language. We report task success and the language ignorance error derived from the conditional mutual information formulation in Eq.~\eqref{eq:language-ignorance-index}.

\textbf{Custom Ambiguity Benchmark (CAB).}
To evaluate ambiguity detection and selective prediction, we design a targeted benchmark with controlled attribute level collisions (see Table~\ref{tab:cab-stats}).

\textit{Setup.}
    Each scene contains several objects of the same type but with distinct attributes, for example two red blocks, one blue block, and one red apple. Correct grounding, therefore, requires attribute-level discrimination rather than type-only matching.

\textit{Instruction types.}
    We define two categories: \emph{Unambiguous:} for example, ``pick up the blue block'', ``get the apple'' and
\emph{Ambiguous:} for example, ``pick up the block'', ``get the red one'', where multiple entities satisfy the description.

\textit{Success criteria.}
    A model succeeds if:
    (a) it correctly executes unambiguous instructions, and
    (b) for ambiguous instructions, it abstains by outputting a clarification token (``[CLARIFY]'') instead of committing to an arbitrary entity.
    Acting under ambiguity is counted as failure. We also measure SACA attention entropy (Eq.~\eqref{eq:saca-entropy}) and analyze its relation to ambiguity through the entropy-based selective policy and its induced Risk Coverage curve.

\begin{table}[t]
\centering
\footnotesize
\caption{Summary of Custom Ambiguity Benchmark.}
\label{tab:cab-stats}
\begin{tabular}{l c}
\toprule
\textbf{Quantity} & \textbf{Value} \\
\midrule
Scenes (train/val/test) & $32 / 8 / 8$ \\
Objects per scene & $3 / 4.8 / 6$ \\
(min / mean / max) & \\

Object categories & blocks, mugs, \\
& bottles, fruit \\

Attributes & colors: 4 $\approx25\%$ each; \\
& sizes: 2 $\approx50\%$ each \\

Instructions & $2{,}400$ total \\

Unambig / Ambig & $1{,}200 / 1{,}200$ \\

Ambiguity rule & drop attrs until $\ge2$ \\
& entities match \\

Unambig success & correct pick, no clarify \\

Ambig success & clarify issued before pick \\

Clarify module & entropy gated \texttt{[CLARIFY]} \\

Eval split & metrics on held out test scenes \\
\bottomrule\vspace{-20pt}
\end{tabular}
\end{table}

\subsection{Ablation Studies}

We perform controlled ablations to isolate the contribution of each component relative to the full ProGAL-VLA model. This ablation shows that the planner’s contribution is largely lexical regularization rather than additional model capacity; when replaced with a non-LLM template extractor, most grounding and robustness gains remain. 

\textbf{Base VLA (OpenVLA).}
This is the original OpenVLA model fine-tuned on LIBERO-Plus~\cite{fei2025libero} without any grounding or ambiguity mechanisms. It serves as the primary baseline.

\textbf{ProGAL-VLA without $\pi_{\text{slow}}$.}
To test the role of hierarchical planning, we replace the sub-goal query with a direct instruction query by setting
$Q = \mathrm{Embed}_{\text{sym}}(L)$
instead of
$Q = \mathrm{Embed}_{\text{sym}}(s_t)$.
This evaluates whether a single-stage VLA with entity-aware attention can match the planner variant.

\textbf{ProGAL-VLA without GSM.}
To assess explicit 3D entity-centric state, we remove GSM Graph and GSM Memory and feed SACA with keys and values derived from generic visual features (for example patch tokens). This examines whether structured object-level representations are necessary for stable grounding and ambiguity awareness.

\textbf{ProGAL-VLA without $\mathcal{L}_{\text{GAC}}$.}
We train the full architecture using only the action loss $\mathcal{L}_{\text{action}}$, omitting the contrastive grounding objective. This isolates the influence of the entity level mutual information lower bound in Eq.~\eqref{eq:info-bound-gac} on retrieval, language sensitivity, and ambiguity detection.

Across LIBERO-Plus, robustness tests, and CAB, the complete ProGAL-VLA model consistently outperforms all ablations. The largest gains appear in language ignorance, entity retrieval, and ambiguity metrics, confirming that hierarchical planning, structured GSM state, and the GAC objective all contribute materially to the final performance.

\begin{table*}[t]
\centering
\caption{Robustness across perturbation dimensions on LIBERO-Plus~\cite{fei2025libero}. Bold values mark the best score in each column. The bottom row reports ProGAL-VLA along with absolute gains over the base OpenVLA model.}
\resizebox{\textwidth}{!}{
\begin{tabular}{lccccccc|c}
\toprule
\textbf{Model} & \textbf{Camera} & \textbf{Robot} & \textbf{Language} & \textbf{Light} & \textbf{Background} & \textbf{Noise} & \textbf{Layout} & \textbf{Total} \\
\midrule
OpenVLA~\cite{kim2024openvla} & 0.8 & 3.5 & 23.0 & 8.1 & 34.8 & 15.2 & 28.5 & 17.3 \\
OpenVLA-OFT~\cite{kim2025fine} & 56.4 & 31.9 & 79.5 & 88.7 & 93.3 & 75.8 & 74.2 & 70.0 \\
OpenVLA-OFT\_w~\cite{kim2025fine} & 10.4 & 38.7 & 70.5 & 76.8 & 93.6 & 49.9 & 69.9 & 56.4 \\
NORA~\cite{hung2025norasmallopensourcedgeneralist} & 2.2 & 37.0 & 65.1 & 45.7 & 58.6 & 12.8 & 62.1 & 39.8 \\
WorldVLA~\cite{cen2025worldvla} & 0.1 & 27.9 & 41.6 & 43.7 & 17.1 & 10.9 & 38.0 & 25.3 \\
UniVLA~\cite{bu2025univla} & 1.8 & 46.2 & 69.6 & 69.0 & 81.0 & 21.2 & 31.9 & 43.9 \\
$\pi_0$~\cite{black2410pi0} & 13.8 & 6.0 & 58.8 & 85.0 & 81.4 & 79.0 & 68.9 & 54.6 \\
$\pi_0$-Fast~\cite{pertsch2025fast} & 65.1 & 21.6 & 61.0 & 73.2 & 73.2 & 74.4 & 68.8 & 64.2 \\
RIPT-VLA~\cite{tan2025interactiveposttrainingvisionlanguageactionmodels} & 55.2 & 31.2 & 77.6 & 88.4 & 91.6 & 73.5 & 74.2 & 69.3 \\
OpenVLA-OFT\_m~\cite{kim2025fine} & 55.6 & 21.7 & 81.0 & 92.7 & 91.0 & 78.6 & 68.7 & 68.1 \\
OpenVLA-OFT+~\cite{kim2025fine} & 92.8 & 30.3 & 85.8 & \textbf{94.9} & \textbf{93.9} & \textbf{89.3} & 77.6 & 79.6 \\
ProGAL-VLA (Ours) & \textbf{93.2} & \textbf{71.5} & \textbf{93.6} & 86.8 & 92.3 & 74.8 & \textbf{86.7} & \textbf{85.5} \\
\bottomrule\vspace{-20pt}
\end{tabular}
}
\label{tab:libero-robustness}
\end{table*}

\begin{figure}[t]
    \centering
    \includegraphics[width=\linewidth]{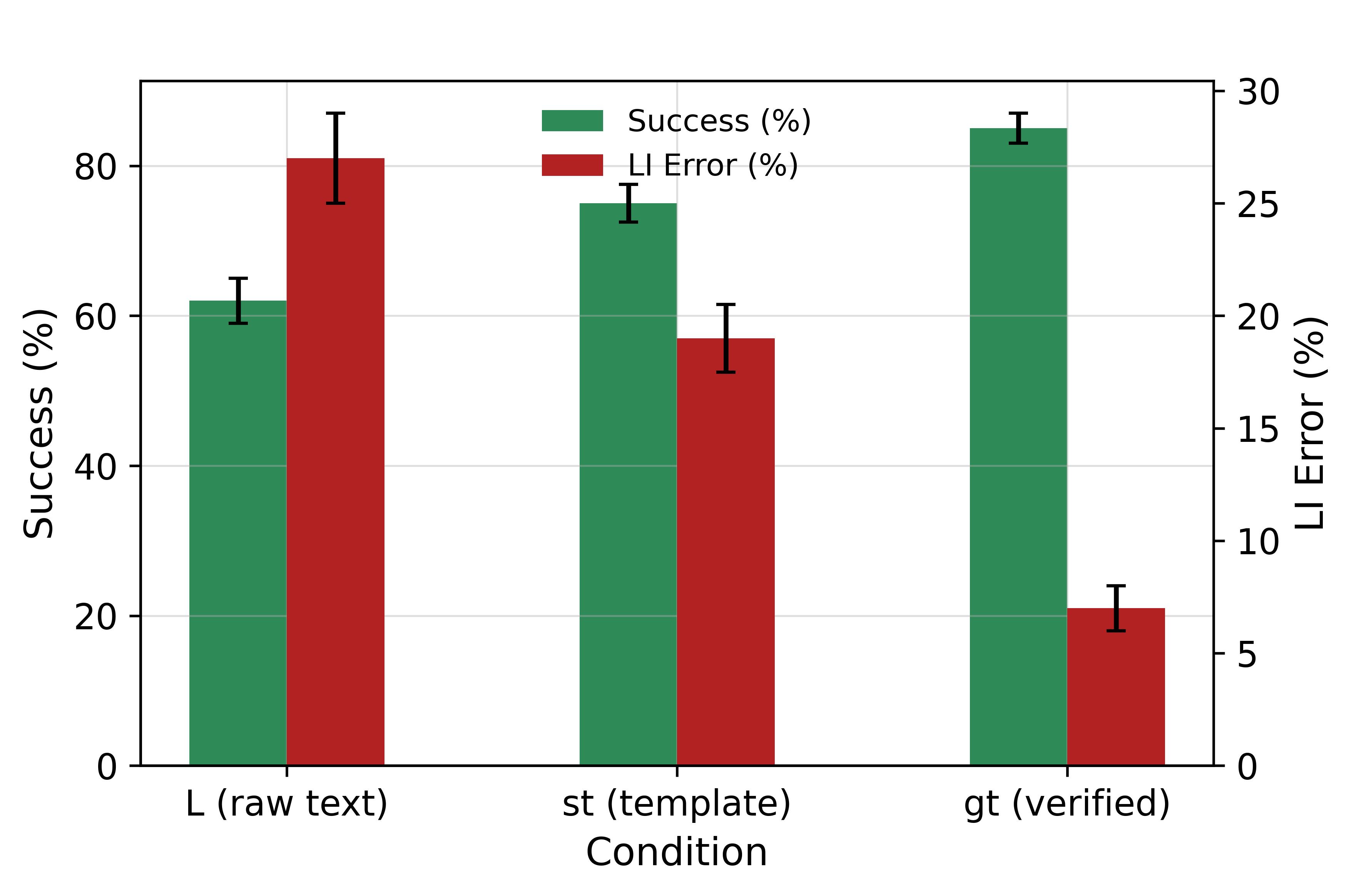}
    \caption{Success rate and language-ignorance error on LIBERO-Plus~\cite{fei2025libero} across different input configurations ($L$, $s_t$, and $g_t$).
    Conditioning on the verified grounding $g_t$ yields both higher task success and lower language-ignorance error, consistent with the analysis in Sec.~\ref{sec:theory}.}
    \label{fig:success_LI}
\end{figure}

\begin{figure}[t]
    \centering
    \includegraphics[width=\linewidth]{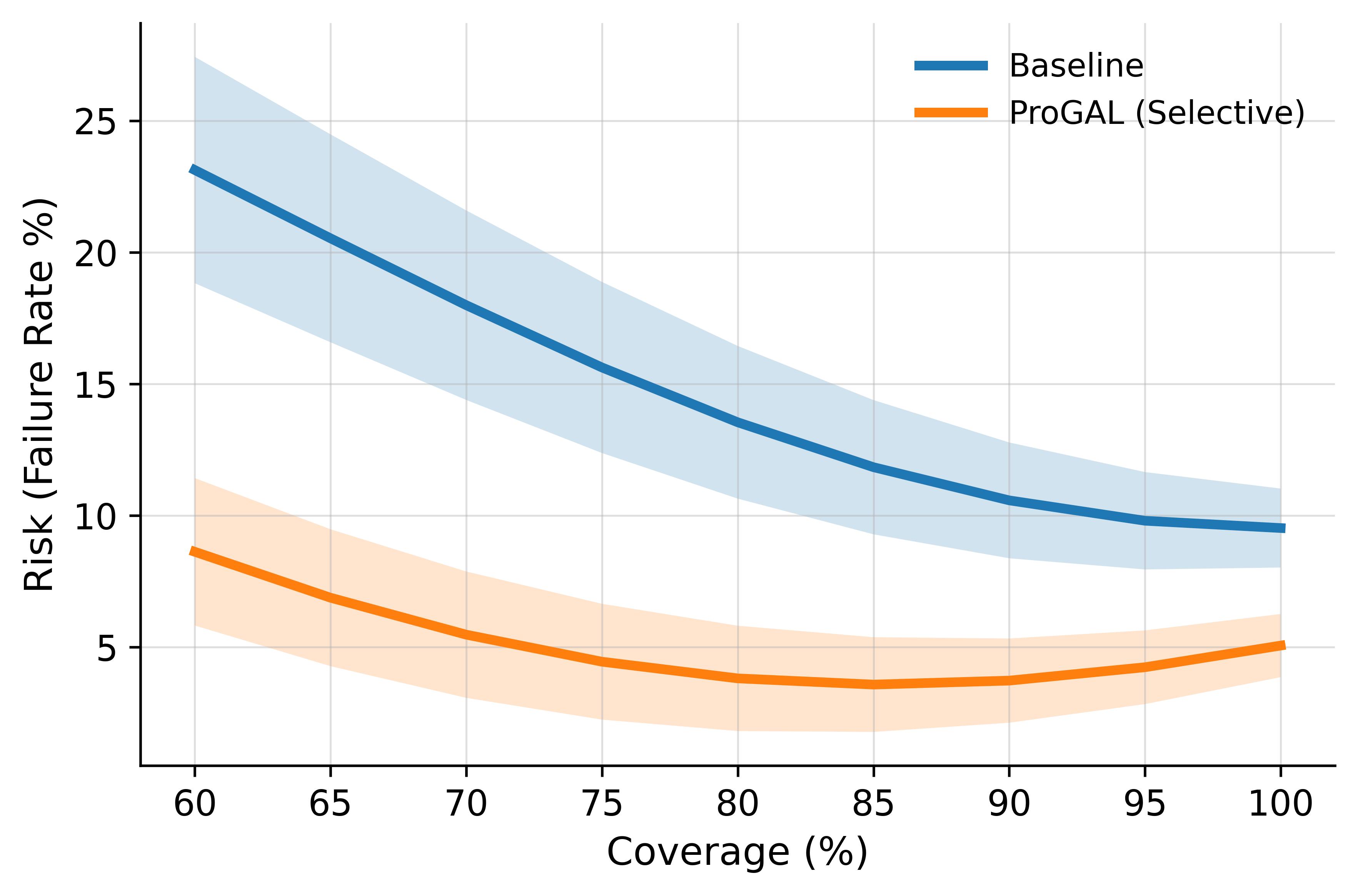}
\caption{Selective prediction performance on CAB. Risk-Coverage curves from entropy-based selective policies show that ProGAL-VLA maintains strictly lower failure risk across all coverage levels, indicating better calibrated uncertainty and more reliable abstention on ambiguous inputs.}

    \label{fig:risk_coverage}
\end{figure}

\begin{table*}[t]
\centering
\caption{Ambiguity detection and selective prediction on CAB. Bold entries mark the best performance. AUROC, AUPR, Cov@95, Clar@Ambig, and Unambig SR are higher is better; ECE and FPR@95 are lower is better. Total is the macro average of all normalized ambiguity metrics.}
\resizebox{\textwidth}{!}{
\begin{tabular}{lccccccc|c}
\toprule
\textbf{Model} & \textbf{AUROC}$\uparrow$ & \textbf{AUPR}$\uparrow$ & \textbf{ECE}$\downarrow$ & \textbf{Cov@95}$\uparrow$ & \textbf{FPR@95}$\downarrow$ & \textbf{Clar@Ambig}$\uparrow$ & \textbf{Unambig SR}$\uparrow$ & \textbf{Total}$\uparrow$ \\
\midrule
OpenVLA~\cite{kim2024openvla} & 0.52 & 0.49 & 14.3 & 0.31 & 0.72 & 0.09 & 0.74 & 0.47 \\
ProGAL (w/o $\mathcal{L}_{\text{GAC}}$) & 0.66 & 0.63 & 9.8 & 0.57 & 0.41 & 0.42 & 0.82 & 0.64 \\
ProGAL (entropy-only) & 0.73 & 0.70 & 7.1 & 0.63 & 0.33 & 0.58 & 0.84 & 0.69 \\
\textbf{ProGAL-VLA (Ours)} & \textbf{0.81} & \textbf{0.79} & \textbf{4.6} & \textbf{0.78} & \textbf{0.18} & \textbf{0.81} & \textbf{0.89} & \textbf{0.79} \\
\bottomrule\vspace{-20pt}
\end{tabular}
}
\label{tab:cab-ambiguity}
\end{table*}

\begin{table*}[t]
\centering
\caption{Summary of entity retrieval from SACA logits and language-ignorance (lower is better).
For retrieval, we report Recall@1 with varying candidate set sizes.}
\resizebox{\textwidth}{!}{
\begin{tabular}{lcccccccc}
\toprule
& \multicolumn{4}{c}{\textbf{Retrieval (SACA logits)}} 
& \multicolumn{3}{c}{\textbf{Language-ignorance}} \\
\cmidrule(lr){2-5}\cmidrule(lr){6-8}
\textbf{Model} & N=8 R@1$\uparrow$ & N=8 R@5$\uparrow$ & N=16 R@1$\uparrow$ & N=32 R@1$\uparrow$ & Simple$\downarrow$ & Spatial$\downarrow$ & Relational$\downarrow$ \\
\midrule
OpenVLA~\cite{kim2024openvla} & 0.41 & 0.72 & 0.28 & 0.15 & 0.36 & 0.49 & 0.57 \\
ProGAL (w/o $\mathcal{L}_{\text{GAC}}$) 
& 0.57 & 0.86 & 0.43 & 0.29 & 0.22 & 0.31 & 0.45 \\
ProGAL ($L \!\rightarrow\! \pi_{\text{fast}}$) 
& 0.50 & 0.83 & 0.35 & 0.24 & 0.27 & 0.33 & 0.42 \\
\textbf{ProGAL (Ours)} 
& \textbf{0.71} & \textbf{0.93} & \textbf{0.58} & \textbf{0.41} & \textbf{0.08} & \textbf{0.14} & \textbf{0.19} \\
\bottomrule\vspace{-20pt}
\end{tabular}
}
\label{core}
\end{table*}

\begin{table}[t]
\centering
\caption{Per step latency of ProGAL-VLA on LIBERO-Plus~\cite{fei2025libero} over 100 episodes. Values are mean and standard deviation in milliseconds. Throughput: 10.31 FPS.}
\resizebox{\columnwidth}{!}{
\begin{tabular}{lccccc}
\toprule
Detector & GSM & SACA & $\pi_{\text{fast}}$ & Total \\
\midrule
\shortstack[c]{43.0\\ {\scriptsize$\pm$ 26.4}} &
\shortstack[c]{15.8\\ {\scriptsize$\pm$ 9.7}} &
\shortstack[c]{10.7\\ {\scriptsize$\pm$ 6.6}} &
\shortstack[c]{26.9\\ {\scriptsize$\pm$ 16.5}} &
\shortstack[c]{\textbf{96.4}\\ {\scriptsize$\pm$ \textbf{59.2}}} \\
\bottomrule\vspace{-20pt}
\end{tabular}
}
\label{tab:latency-singlecol}
\end{table}

\subsection{Results and Analysis}
\label{sec:results}

\textbf{Robustness on LIBERO-Plus.}
Table~\ref{tab:libero-robustness} shows success rates under seven perturbation dimensions on LIBERO-Plus. OpenVLA is highly brittle, with performance collapsing under camera and robot shifts (for example, 0.8 percent under camera changes). Recent robust variants such as RIPT VLA and OpenVLA OFT+ substantially improve stability but remain uneven across dimensions; OpenVLA OFT+ is strong on lighting, background, and noise, yet comparatively weak under robot and layout perturbations. ProGAL-VLA attains the best score in six of eight columns, with especially large gains under robot shifts (30.3\% to 71.5\%) and layout changes (77.6\% to 86.7\%). Its total score of 85.5\% indicates that verified grounding and the GSM based entity representation yield consistent robustness across heterogeneous nuisance factors rather than overfitting to any single perturbation. This pattern matches the Lipschitz style robustness argument in Eq.~\eqref{eq:robustness-tight}, where a stable verified embedding $g_t$ supports a stable action distribution under camera, lighting, and layout variation.

\textbf{Language ignorance on LIBERO-Plus.}
Figure~\ref{fig:success_LI} compares success rate and language ignorance when the fast policy is conditioned on language $L$, planner sub-goals $s_t$, or the verified embedding $g_t$. Conditioning on $L$ reproduces the language ignorance failure: success improves over observation-only baselines, but actions change little under instruction perturbations. Using $s_t$ reduces this effect slightly, yet remains limited because sub-goals are not entity-bound. Conditioning on $g_t$ yields both the highest success rate and the lowest language ignorance, consistent with Eq.~\eqref{eq:language-influence-decomposition}: the verified embedding captures information about $L$ beyond $(O_t, q_t)$, increasing $I(L; a_t \mid O_t, q_t)$ and reducing the normalized index $\Lambda^{\pi}$. In practice, verified grounding ensures that instructions exert a clear and measurable influence on actions.

\textbf{Ambiguity detection and selective prediction on CAB.}
Table~\ref{tab:cab-stats} summarizes the statistics and evaluation protocol of the Custom Ambiguity Benchmark (CAB), including scene counts, object and attribute distributions, instruction split, clarification mechanism, and test setup.
Table~\ref{tab:cab-ambiguity} summarizes ambiguity detection and selective prediction results on CAB. ProGAL-VLA surpasses both OpenVLA and the no-$\mathcal{L}_{\text{GAC}}$ variant across all metrics: AUROC increases from 0.52 and 0.66 to 0.81, AUPR from 0.49 and 0.63 to 0.79, and ECE decreases from 14.3 to 4.6. Coverage at 95 percent confidence rises from 0.31 to 0.78, while FPR@95 drops from 0.72 to 0.18, indicating far fewer false safe decisions. Clarification behavior improves accordingly, with Clar@Ambig increasing from 0.09 to 0.81 and unambiguous success improving from 0.74 to 0.89. The overall Total score increases from 0.47 to 0.79. These results show that SACA entropy yields a strong and well-calibrated ambiguity signal, aligning with the selective prediction formulation and the Risk Coverage curve $\mathcal{C}$. Figure~\ref{fig:risk_coverage} illustrates this, with ProGAL-VLA maintaining consistently lower failure risk across the coverage range. CAB targets attribute-level ambiguity, where VLA grounding failures are most pronounced and easiest to quantify. It does not address broader linguistic ambiguity (e.g., pragmatic, temporal, or multi-step references), as the benchmark isolates a minimal, measurable setting; extending to richer ambiguity types is left for future work.

\textbf{Entity retrieval and language ignorance.}
Table~\ref{core} reports retrieval accuracy from SACA logits and language ignorance across instruction types. ProGAL-VLA yields the strongest retrieval at all candidate set sizes, raising Recall@1 over OpenVLA from 0.41 to 0.71 for $N=8$, 0.28 to 0.58 for $N=16$, and 0.15 to 0.41 for $N=32$. Removing $\mathcal{L}_{\text{GAC}}$ or the verified bottleneck degrades performance, with both ablations landing between the extremes. This confirms that the GSM structure and the contrastive grounding loss jointly enable discriminative entity alignment, consistent with the InfoNCE bound in Eq.~\eqref{eq:info-bound-gac}, where optimizing $\mathcal{L}_{\text{GAC}}$ increases $I(S,E)$ and sharpens SACA logits.

Language ignorance shows a similar pattern. OpenVLA yields high scores across simple, spatial, and relational instructions $(0.36, 0.49, 0.57)$. Removing $\mathcal{L}_{\text{GAC}}$ reduces these to $(0.22, 0.31, 0.45)$, and feeding language directly to $\pi_{\text{fast}}$ gives $(0.27, 0.33, 0.42)$. ProGAL-VLA obtains the lowest values $(0.08, 0.14, 0.19)$, showing that combination of hierarchical planning, GSM, SACA, and GAC is essential for ensuring that language meaningfully modulates action for spatial and relational tasks.

\textbf{Runtime and latency.}
Table~\ref{tab:latency-singlecol} reports per-step latency on LIBERO-Plus across the detector, GSM, SACA, and $\pi_{\text{fast}}$. The total end-to-end latency of $\approx 107.5 \pm 66.0$ ms (9.31 FPS) is practical for closed-loop manipulation. Most cost arises from the detector and $\pi_{\text{fast}}$, while GSM and SACA add modest overhead relative to the underlying vision and language backbones. Thus, ProGAL-VLA achieves improved robustness, grounding, and ambiguity handling with a small computational premium.

\section{Conclusion}
\label{sec:conclusion}

We presented ProGAL-VLA, a grounding-aware VLA architecture that separates slow symbolic planning from fast control through an explicit verification bottleneck. The system constructs a 3D entity graph, aligns symbolic sub-goals with grounded entities via the GAC objective, and executes actions only through a verified goal embedding. On LIBERO-Plus and our Custom Ambiguity Benchmark, ProGAL-VLA achieves state-of-the-art robustness, reduces language ignorance, and provides calibrated ambiguity detection. Analysis shows that the verification bottleneck increases conditional mutual information between language and actions, the GAC loss yields an entity-level InfoNCE bound, and SACA entropy enables principled selective prediction. Empirically, these mechanisms improve robustness to camera, layout, and lighting shifts, sharpen entity retrieval, and support reliable abstention under ambiguity. Verified grounding therefore offers a practical route to more instruction-sensitive and stable robotic agents.

\textbf{Limitations.} Our evaluation uses simulated LIBERO-Plus environments and a controlled ambiguity benchmark, which fail to fully capture real scene variability, including occlusions and depth noise. Transferring GSM, SACA, and entropy-based selective policies to physical robots with imperfect sensing is still unresolved. The method relies on pretrained components like YOLO World and large VL backbones, whose errors can lead to grounding failures. The ambiguity benchmark emphasizes attribute collisions and a single clarify action, but real-world scenarios may require more complex interactions. Additionally, the architecture adds system complexity, necessitating further optimization for larger backbones or higher resolution inputs. Despite these challenges, the per-step performance remains.

\section*{Acknowledgment}

This work was supported in part by COGNISENSE, one of seven centers in JUMP 2.0, a Semiconductor Research Corporation (SRC) program sponsored by DARPA and by NSF Award $\#2046435$.

{
    \small
    \bibliographystyle{ieeenat_fullname}
    \bibliography{main}

@misc{li2025_250815201,
      title={Survey of Vision-Language-Action Models for Embodied Manipulation},
      author={Haoran Li and Yuhui Chen and Wenbo Cui and Weiheng Liu and Kai Liu and Mingcai Zhou and Zhengtao Zhang and Dongbin Zhao},
      year={2025},
      eprint={2508.15201},
      archivePrefix={arXiv},
      primaryClass={cs.RO},
      url={http://arxiv.org/abs/2508.15201v2},
}

@article{brohan2022rt,
  title={Rt-1: Robotics transformer for real-world control at scale},
  author={Brohan, Anthony and Brown, Noah and Carbajal, Justice and Chebotar, Yevgen and Dabis, Joseph and Finn, Chelsea and Gopalakrishnan, Keerthana and Hausman, Karol and Herzog, Alex and Hsu, Jasmine and others},
  journal={arXiv preprint arXiv:2212.06817},
  year={2022}
}

@inproceedings{zitkovich2023rt,
  title={Rt-2: Vision-language-action models transfer web knowledge to robotic control},
  author={Zitkovich, Brianna and Yu, Tianhe and Xu, Sichun and Xu, Peng and Xiao, Ted and Xia, Fei and Wu, Jialin and Wohlhart, Paul and Welker, Stefan and Wahid, Ayzaan and others},
  booktitle={Conference on Robot Learning},
  pages={2165--2183},
  year={2023},
  organization={PMLR}
}

@article{driess2023palm,
  title={Palm-e: An embodied multimodal language model},
  author={Driess, Danny and Xia, Fei and Sajjadi, Mehdi SM and Lynch, Corey and Chowdhery, Aakanksha and Wahid, Ayzaan and Tompson, Jonathan and Vuong, Quan and Yu, Tianhe and Huang, Wenlong and others},
  year={2023}
}

@article{reed2022generalist,
  title={A generalist agent},
  author={Reed, Scott and Zolna, Konrad and Parisotto, Emilio and Colmenarejo, Sergio Gomez and Novikov, Alexander and Barth-Maron, Gabriel and Gimenez, Mai and Sulsky, Yury and Kay, Jackie and Springenberg, Jost Tobias and others},
  journal={arXiv preprint arXiv:2205.06175},
  year={2022}
}

@article{kim2024openvla,
  title={Openvla: An open-source vision-language-action model},
  author={Kim, Moo Jin and Pertsch, Karl and Karamcheti, Siddharth and Xiao, Ted and Balakrishna, Ashwin and Nair, Suraj and Rafailov, Rafael and Foster, Ethan and Lam, Grace and Sanketi, Pannag and others},
  journal={arXiv preprint arXiv:2406.09246},
  year={2024}
}

@article{balachandran2024eureka,
  title={Eureka: Evaluating and understanding large foundation models},
  author={Balachandran, Vidhisha and Chen, Jingya and Joshi, Neel and Nushi, Besmira and Palangi, Hamid and Salinas, Eduardo and Vineet, Vibhav and Woffinden-Luey, James and Yousefi, Safoora},
  journal={arXiv preprint arXiv:2409.10566},
  year={2024}
}

@article{intelligence2025pi_,
  title={$\pi_{0.5}$: a Vision-Language-Action Model with Open-World Generalization},
  author={Intelligence, Physical and Black, Kevin and Brown, Noah and Darpinian, James and Dhabalia, Karan and Driess, Danny and Esmail, Adnan and Equi, Michael and Finn, Chelsea and Fusai, Niccolo and others},
  journal={arXiv preprint arXiv:2504.16054},
  year={2025}
}

@article{fei2025libero,
  title={LIBERO-Plus: In-depth Robustness Analysis of Vision-Language-Action Models},
  author={Fei, Senyu and Wang, Siyin and Shi, Junhao and Dai, Zihao and Cai, Jikun and Qian, Pengfang and Ji, Li and He, Xinzhe and Zhang, Shiduo and Fei, Zhaoye and others},
  journal={arXiv preprint arXiv:2510.13626},
  year={2025}
}

@misc{zhu2024_241003659,
      title={Unraveling Cross-Modality Knowledge Conflicts in Large Vision-Language Models},
      author={Tinghui Zhu and Qin Liu and Fei Wang and Zhengzhong Tu and Muhao Chen},
      year={2024},
      eprint={2410.03659},
      archivePrefix={arXiv},
      primaryClass={cs.CV},
      url={http://arxiv.org/abs/2410.03659v2},
}

@misc{lu2023_230904041,
      title={Evaluation and Enhancement of Semantic Grounding in Large Vision-Language Models},
      author={Jiaying Lu and Jinmeng Rao and Kezhen Chen and Xiaoyuan Guo and Yawen Zhang and Baochen Sun and Carl Yang and Jie Yang},
      year={2023},
      eprint={2309.04041},
      archivePrefix={arXiv},
      primaryClass={cs.CV},
      url={http://arxiv.org/abs/2309.04041v2},
}

@misc{han2025_250402477,
      title={Multimodal Fusion and Vision-Language Models: A Survey for Robot Vision},
      author={Xiaofeng Han and Shunpeng Chen and Zenghuang Fu and Zhe Feng and Lue Fan and Dong An and Changwei Wang and Li Guo and Weiliang Meng and Xiaopeng Zhang and Rongtao Xu and Shibiao Xu},
      year={2025},
      eprint={2504.02477},
      archivePrefix={arXiv},
      primaryClass={cs.RO},
      url={http://arxiv.org/abs/2504.02477v3},
      doi={10.1016/j.inffus.2025.103652},
}

@misc{toyoda2021_210408521,
      title={Embodying Pre-Trained Word Embeddings Through Robot Actions},
      author={Minori Toyoda and Kanata Suzuki and Hiroki Mori and Yoshihiko Hayashi and Tetsuya Ogata},
      year={2021},
      eprint={2104.08521},
      archivePrefix={arXiv},
      primaryClass={cs.RO},
      url={http://arxiv.org/abs/2104.08521v1},
      doi={10.1109/LRA.2021.3067862},
}

@article{shi2025memoryvla,
  title={Memoryvla: Perceptual-cognitive memory in vision-language-action models for robotic manipulation},
  author={Shi, Hao and Xie, Bin and Liu, Yingfei and Sun, Lin and Liu, Fengrong and Wang, Tiancai and Zhou, Erjin and Fan, Haoqiang and Zhang, Xiangyu and Huang, Gao},
  journal={arXiv preprint arXiv:2508.19236},
  year={2025}
}

@article{zhang2025pure,
  title={Pure Vision Language Action (VLA) Models: A Comprehensive Survey},
  author={Zhang, Dapeng and Sun, Jin and Hu, Chenghui and Wu, Xiaoyan and Yuan, Zhenlong and Zhou, Rui and Shen, Fei and Zhou, Qingguo},
  journal={arXiv preprint arXiv:2509.19012},
  year={2025}
}

@article{huang2025graphcot,
  title={Graphcot-vla: A 3d spatial-aware reasoning vision-language-action model for robotic manipulation with ambiguous instructions},
  author={Huang, Helong and Cen, Min and Tan, Kai and Quan, Xingyue and Huang, Guowei and Zhang, Hong},
  journal={arXiv preprint arXiv:2508.07650},
  year={2025}
}

@article{chang2021comprehensive,
  title={A comprehensive survey of scene graphs: Generation and application},
  author={Chang, Xiaojun and Ren, Pengzhen and Xu, Pengfei and Li, Zhihui and Chen, Xiaojiang and Hauptmann, Alex},
  journal={IEEE Transactions on Pattern Analysis and Machine Intelligence},
  volume={45},
  number={1},
  pages={1--26},
  year={2021},
  publisher={IEEE}
}

@inproceedings{mavridis2006grounded,
  title={Grounded situation models for robots: Where words and percepts meet},
  author={Mavridis, Nikolaos and Roy, Deb},
  booktitle={2006 IEEE/RSJ international conference on intelligent robots and systems},
  pages={4690--4697},
  year={2006},
  organization={IEEE}
}

@inproceedings{brohan2023can,
  title={Do as i can, not as i say: Grounding language in robotic affordances},
  author={Brohan, Anthony and Chebotar, Yevgen and Finn, Chelsea and Hausman, Karol and Herzog, Alexander and Ho, Daniel and Ibarz, Julian and Irpan, Alex and Jang, Eric and Julian, Ryan and others},
  booktitle={Conference on robot learning},
  pages={287--318},
  year={2023},
  organization={PMLR}
}

@inproceedings{li2024gp,
  title={Gp-nerf: Generalized perception nerf for context-aware 3d scene understanding},
  author={Li, Hao and Zhang, Dingwen and Dai, Yalun and Liu, Nian and Cheng, Lechao and Li, Jingfeng and Wang, Jingdong and Han, Junwei},
  booktitle={Proceedings of the IEEE/CVF conference on computer vision and pattern recognition},
  pages={21708--21718},
  year={2024}
}

@article{maggio2022loc,
  title={Loc-nerf: Monte carlo localization using neural radiance fields},
  author={Maggio, Dominic and Abate, Marcus and Shi, Jingnan and Mario, Courtney and Carlone, Luca},
  journal={arXiv preprint arXiv:2209.09050},
  year={2022}
}

@article{liang2022code,
  title={Code as policies: Language model programs for embodied control},
  author={Liang, Jacky and Huang, Wenlong and Xia, Fei and Xu, Peng and Hausman, Karol and Ichter, Brian and Florence, Pete and Zeng, Andy},
  journal={arXiv preprint arXiv:2209.07753},
  year={2022}
}

@article{black2410pi0,
  title={$\pi$0: A vision-language-action flow model for general robot control. CoRR, abs/2410.24164, 2024. doi: 10.48550},
  author={Black, Kevin and Brown, Noah and Driess, Danny and Esmail, Adnan and Equi, Michael and Finn, Chelsea and Fusai, Niccolo and Groom, Lachy and Hausman, Karol and Ichter, Brian and others},
  journal={arXiv preprint ARXIV.2410.24164}
}

@article{ma2024survey,
  title={A survey on vision-language-action models for embodied ai},
  author={Ma, Yueen and Song, Zixing and Zhuang, Yuzheng and Hao, Jianye and King, Irwin},
  journal={arXiv preprint arXiv:2405.14093},
  year={2024}
}

@article{pertsch2025fast,
  title={Fast: Efficient action tokenization for vision-language-action models},
  author={Pertsch, Karl and Stachowicz, Kyle and Ichter, Brian and Driess, Danny and Nair, Suraj and Vuong, Quan and Mees, Oier and Finn, Chelsea and Levine, Sergey},
  journal={arXiv preprint arXiv:2501.09747},
  year={2025}
}

@misc{tan2025interactiveposttrainingvisionlanguageactionmodels,
      title={Interactive Post-Training for Vision-Language-Action Models}, 
      author={Shuhan Tan and Kairan Dou and Yue Zhao and Philipp Krähenbühl},
      year={2025},
      eprint={2505.17016},
      archivePrefix={arXiv},
      primaryClass={cs.LG},
      url={https://arxiv.org/abs/2505.17016}, 
}

@article{bu2025univla,
  title={Univla: Learning to act anywhere with task-centric latent actions},
  author={Bu, Qingwen and Yang, Yanting and Cai, Jisong and Gao, Shenyuan and Ren, Guanghui and Yao, Maoqing and Luo, Ping and Li, Hongyang},
  journal={arXiv preprint arXiv:2505.06111},
  year={2025}
}

@article{cen2025worldvla,
  title={WorldVLA: Towards Autoregressive Action World Model},
  author={Cen, Jun and Yu, Chaohui and Yuan, Hangjie and Jiang, Yuming and Huang, Siteng and Guo, Jiayan and Li, Xin and Song, Yibing and Luo, Hao and Wang, Fan and others},
  journal={arXiv preprint arXiv:2506.21539},
  year={2025}
}

@misc{hung2025norasmallopensourcedgeneralist,
      title={NORA: A Small Open-Sourced Generalist Vision Language Action Model for Embodied Tasks}, 
      author={Chia-Yu Hung and Qi Sun and Pengfei Hong and Amir Zadeh and Chuan Li and U-Xuan Tan and Navonil Majumder and Soujanya Poria},
      year={2025},
      eprint={2504.19854},
      archivePrefix={arXiv},
      primaryClass={cs.RO},
      url={https://arxiv.org/abs/2504.19854}, 
}

@inproceedings{radford2021learning,
  title={Learning transferable visual models from natural language supervision},
  author={Radford, Alec and Kim, Jong Wook and Hallacy, Chris and Ramesh, Aditya and Goh, Gabriel and Agarwal, Sandhini and Sastry, Girish and Askell, Amanda and Mishkin, Pamela and Clark, Jack and others},
  booktitle={International conference on machine learning},
  pages={8748--8763},
  year={2021},
  organization={PmLR}
}

@article{kim2025fine,
  title={Fine-Tuning Vision-Language-Action Models: Optimizing Speed and Success},
  author={Kim, Moo Jin and Finn, Chelsea and Liang, Percy},
  journal={arXiv preprint arXiv:2502.19645},
  year={2025}
}

@article{qian20243d,
  title={3d-mvp: 3d multiview pretraining for robotic manipulation},
  author={Qian, Shengyi and Mo, Kaichun and Blukis, Valts and Fouhey, David F and Fox, Dieter and Goyal, Ankit},
  journal={arXiv preprint arXiv:2406.18158},
  year={2024}
}

@inproceedings{radosavovic2023real,
  title={Real-world robot learning with masked visual pre-training},
  author={Radosavovic, Ilija and Xiao, Tete and James, Stephen and Abbeel, Pieter and Malik, Jitendra and Darrell, Trevor},
  booktitle={Conference on Robot Learning},
  pages={416--426},
  year={2023},
  organization={PMLR}
}

@article{nair2022r3m,
  title={R3m: A universal visual representation for robot manipulation},
  author={Nair, Suraj and Rajeswaran, Aravind and Kumar, Vikash and Finn, Chelsea and Gupta, Abhinav},
  journal={arXiv preprint arXiv:2203.12601},
  year={2022}
}

@misc{puertamerino2025_250108068,
      title={A Roadmap to Guide the Integration of LLMs in Hierarchical Planning},
      author={Israel Puerta-Merino and Carlos Núñez-Molina and Pablo Mesejo and Juan Fernández-Olivares},
      year={2025},
      eprint={2501.08068},
      archivePrefix={arXiv},
      primaryClass={cs.AI},
      url={http://arxiv.org/abs/2501.08068v1},
}

@misc{han2024_240414285,
      title={LLM-Personalize: Aligning LLM Planners with Human Preferences via Reinforced Self-Training for Housekeeping Robots},
      author={Dongge Han and Trevor McInroe and Adam Jelley and Stefano V. Albrecht and Peter Bell and Amos Storkey},
      year={2024},
      eprint={2404.14285},
      archivePrefix={arXiv},
      primaryClass={cs.RO},
      url={http://arxiv.org/abs/2404.14285v3},
}
}

\clearpage
\setcounter{page}{1}
\maketitlesupplementary

\section{Theoretical Derivations and Proofs}
\label{sec:proofs}

This section expands the theoretical analysis from Section 4 and provides fully detailed derivations for the results used in the main paper.

\subsection{Proof of Proposition 1 (Language Influence)}

\noindent\textbf{Proposition 1.}
Under the Verification Bottleneck assumption,
\begin{equation}
I(L; a_t \mid O_t, q_t)
=
I(L; g_t \mid O_t, q_t)
-
I(L; g_t \mid a_t, O_t, q_t).
\end{equation}

\noindent\textbf{Proof.}

We first introduce a short notation to simplify expressions:
\[
C_t = (O_t, q_t),
\]
which denotes the complete perceptual and robot-state context available at time $t$.

Our goal is to relate the conditional mutual information between the language instruction $L$ and the action $a_t$ to the mutual information involving the verified goal $g_t$.

Consider the joint quantity $I(L; a_t, g_t \mid C_t)$.
The chain rule for mutual information allows two valid decompositions:

\begin{align}
I(L; a_t, g_t \mid C_t)
&= I(L; g_t \mid C_t) + I(L; a_t \mid g_t, C_t),
\label{eq:proof_chain_decomp1}
\\[4pt]
I(L; a_t, g_t \mid C_t)
&= I(L; a_t \mid C_t) + I(L; g_t \mid a_t, C_t).
\label{eq:proof_chain_decomp2}
\end{align}

These two expressions describe the same joint mutual information, but factor it in different orders.

The Verification Bottleneck states that:
\[
a_t \perp L \mid (g_t, C_t).
\]

This means: once the verified goal $g_t$ is known (and the context $C_t$ is fixed), the action no longer depends on the original instruction $L$.  
This architectural constraint is enforced because the fast controller receives only the verified goal $g_t$ and does not have direct access to $L$.

In information-theoretic terms, this conditional independence gives:
\begin{equation}
I(L; a_t \mid g_t, C_t) = 0.
\label{eq:proof_independence}
\end{equation}

Replacing the last term in Eq.~\eqref{eq:proof_chain_decomp1} using Eq.~\eqref{eq:proof_independence}, we obtain:
\[
I(L; a_t, g_t \mid C_t)
=
I(L; g_t \mid C_t).
\]

Both chain-rule decompositions describe the same quantity, so we set  
Eq.~\eqref{eq:proof_chain_decomp1} equal to Eq.~\eqref{eq:proof_chain_decomp2}:

\[
I(L; g_t \mid C_t)
=
I(L; a_t \mid C_t) + I(L; g_t \mid a_t, C_t).
\]

Rearranging the above expression gives:
\[
I(L; a_t \mid C_t)
=
I(L; g_t \mid C_t)
-
I(L; g_t \mid a_t, C_t),
\]
which is exactly the desired identity.

Finally, substituting back $C_t = (O_t, q_t)$ completes the proof.

\subsection{Proof of Theorem 1 (GAC InfoNCE Bound)}

\noindent\textbf{Theorem 1 (InfoNCE style lower bound).}
Assume that, for each symbolic sub goal $s$, one positive grounded entity
$e^+$ is drawn from $p(e \mid s)$ and $N-1$ negatives $\{e_i\}_{i \ne +}$
are drawn independently from the marginal $p(e)$.
Let $\mathcal{L}_{GAC}$ be the contrastive loss
\begin{equation}
\mathcal{L}_{GAC}
=
-\mathbb{E}
\left[
\log
\frac{\exp(f(s, e^+))}
{\sum_{i=1}^{N} \exp(f(s, e_i))}
\right],
\end{equation}
where $f(s,e)$ is a critic function.
Then
\begin{equation}
I(S;E) \;\ge\; \log N \;-\; \mathbb{E}[\mathcal{L}_{GAC}].
\end{equation}

\noindent\textbf{Proof.}
We follow the standard construction used for InfoNCE.
Define a joint distribution over $(S, E_1, \dots, E_N, I)$ as
\begin{equation}
p(s, e_1, \dots, e_N, i)
=
\frac{1}{N}
\, p(s)\, p(e_i \mid s)
\prod_{j \ne i} p(e_j),
\label{eq:info_joint}
\end{equation}
where $I \in \{1,\dots,N\}$ is the index of the positive example.
Under this construction, the conditional posterior of $I$ given
$(s, e_1, \dots, e_N)$ is
\begin{align}
p(i \mid s, e_1,\dots,e_N)
&=
\frac{
p(e_i \mid s) / p(e_i)
}{
\sum_{j=1}^{N} p(e_j \mid s) / p(e_j)
}
\\[4pt]
&=
\frac{
\exp\big(\log p(e_i \mid s) - \log p(e_i)\big)
}{
\sum_{j=1}^{N} 
\exp\big(\log p(e_j \mid s) - \log p(e_j)\big)
}.
\label{eq:true_posterior}
\end{align}

Given a critic $f(s,e)$, we define the model posterior
\begin{equation}
p_f(i \mid s, e_1, \dots, e_N)
=
\frac{\exp(f(s, e_i))}
{\sum_{j=1}^{N} \exp(f(s, e_j))}.
\label{eq:model_posterior}
\end{equation}
The GAC loss can then be written as the expected negative log likelihood
of the true index $I$ under this model:
\begin{equation}
\mathcal{L}_{GAC}
=
\mathbb{E}
\big[
-\log p_f(I \mid S, E_1, \dots, E_N)
\big].
\label{eq:gac_as_ce}
\end{equation}

Now consider the Kullback–Leibler divergence between the true posterior
and the model posterior, conditioned on $(S, E_1, \dots, E_N)$:
\begin{equation}
\mathrm{KL}
\big(
p(\cdot \mid s, e_{1:N})
\;\Vert\;
p_f(\cdot \mid s, e_{1:N})
\big)
\;\ge\; 0.
\end{equation}
Taking expectation over $(S, E_1, \dots, E_N)$ and expanding gives
\begin{align}
0
&\le
\mathbb{E}
\Big[
\mathrm{KL}
\big(
p(\cdot \mid S, E_{1:N})
\;\Vert\;
p_f(\cdot \mid S, E_{1:N})
\big)
\Big]
\\
&=
\mathbb{E}
\big[
-\log p_f(I \mid S, E_{1:N})
\big]
-
\mathbb{E}
\big[
-\log p(I \mid S, E_{1:N})
\big]
\\
&=
\mathbb{E}[\mathcal{L}_{GAC}]
-
\mathbb{E}
\big[
-\log p(I \mid S, E_{1:N})
\big],
\end{align}
where we used Eq.~\eqref{eq:gac_as_ce} in the last line.

Rearranging yields
\begin{equation}
\mathbb{E}[\mathcal{L}_{GAC}]
\;\ge\;
\mathbb{E}
\big[
-\log p(I \mid S, E_{1:N})
\big].
\label{eq:bound_ce}
\end{equation}

The right-hand side can be evaluated in closed form
for the joint in Eq.~\eqref{eq:info_joint}
(see for example Poole et al., 2019).
One obtains
\begin{equation}
\mathbb{E}
\big[
-\log p(I \mid S, E_{1:N})
\big]
=
\log N - I(S;E),
\end{equation}
which substituted into Eq.~\eqref{eq:bound_ce} gives
\begin{equation}
\mathbb{E}[\mathcal{L}_{GAC}]
\;\ge\;
\log N - I(S;E).
\end{equation}
Rearranging terms finally yields
\begin{equation}
I(S;E)
\;\ge\;
\log N - \mathbb{E}[\mathcal{L}_{GAC}],
\end{equation}
That is, the GAC objective defines a variational lower bound
on the mutual information between symbolic sub-goals and grounded entities.

\section{Model Details}
\label{sec:model_details}

This section summarizes the architectures of all policies evaluated on LIBERO-Plus~\cite{fei2025libero}. We focus on the high-level design of the backbones, modality encoders, and action parameterizations.

\noindent\textbf{{OpenVLA and OpenVLA-OFT Family:}} OpenVLA~\cite{kim2024openvla} is built around the Prismatic-7B vision–language backbone. Visual observations are encoded by a dual-stream vision module (SigLIP and DINOv2) whose feature maps are concatenated and passed through a
lightweight MLP projector into the token space of a LLaMA-2–style language model. Text and image tokens are fused via cross-attention, and continuous control commands are discretized into 256 bins per dimension and represented as
dedicated action tokens in the LLM vocabulary, enabling the policy to operate purely in token space.

The OpenVLA-OFT variants~\cite{kim2025fine} share the same multimodal backbone but differ in the action head and input configuration. OpenVLA-OFT replaces the discrete-token head with a continuous MLP head that predicts all action dimensions in parallel and adds FiLM-style conditioning layers to strengthen language grounding. OpenVLA-OFT\_w removes the wrist-camera stream and conditions only on the third-person view, while OpenVLA-OFT\_m is used as a single multi-suite policy that shares parameters across all LIBERO suites. OpenVLA-OFT+~\cite{kim2025fine} further refines this family with an enhanced action head and improved conditioning, while keeping the same core VLM architecture.

\noindent\textbf{{\texorpdfstring{$\pi_0$}{pi0} and $\pi_0$-Fast:}} The $\pi_0$ architecture~\cite{black2410pi0} follows a Transfusion-style design that couples a large multimodal backbone with an action-specialized transformer. A PaliGemma-based VLM encodes multiple RGB views, natural-language
instructions, and proprioceptive state $q_t$ into a token sequence; a subset of these tokens is routed to a smaller ``action expert'' transformer that is responsible for predicting future actions, while the VLM maintains semantic and visual understanding.

$\pi_0$-Fast~\cite{pertsch2025fast} preserves the same backbone but changes the action representation. Instead of working in the raw time domain, trajectories
are transformed into a sparse frequency-domain representation using a discrete cosine transform and then compressed with a BPE-style tokenizer (FAST). This yields shorter discrete action sequences while retaining high-fidelity control.

\noindent\textbf{{NORA:}} NORA~\cite{hung2025norasmallopensourcedgeneralist} is a 3B-parameter VLA model built on the Qwen-2.5-VL-3B multimodal backbone. It processes a natural-language instruction together with a single RGB observation and outputs discrete action tokens using the FAST+ tokenizer, which discretizes continuous actions into a compact token space. The design targets a balance between strong visual–semantic reasoning and practical deployment on moderate hardware.

\noindent\textbf{{WorldVLA:}} WorldVLA~\cite{cen2025worldvla} unifies policy and world modeling in a single autoregressive transformer initialized from Chameleon. All modalities are tokenized: images via a VQ-GAN codebook, text via a BPE tokenizer, and robot
actions via an action tokenizer that maps each control dimension to 256 discrete bins. These tokens are interleaved into one sequence and modeled jointly. A custom attention mask prevents the current action tokens from attending to previous actions, encouraging stable parallel prediction of action ``chunks'' while still conditioning on the full history of images and text.

\noindent\textbf{{UniVLA:}} UniVLA~\cite{bu2025univla} aims for cross-embodiment generalization by operating in a discrete latent action space. The model uses the same Prismatic-7B VLM stack (SigLIP + DINOv2 encoders with a LLaMA-2 backbone) and extends the LLM vocabulary with latent action tokens drawn from a learned codebook. At deployment time, the policy autoregressively predicts sequences of latent tokens from visual observations and instructions. A lightweight robot-specific decoder head, implemented with multi-head attention over the latent token sequence, maps these latents into low-level control commands. UniVLA also supports history-augmented inputs where previously executed latent actions are fed back as context, which is particularly beneficial for long-horizon tasks.

\noindent\textbf{{RIPT-VLA:}} RIPT-VLA~\cite{tan2025interactiveposttrainingvisionlanguageactionmodels} starts
from the continuous-action OpenVLA-OFT policy~\cite{kim2025fine} and augments
it with a distributional action head. The original MLP head produces the mean
$\mu_\theta$ of a factorized Laplace distribution over actions, and an
additional lightweight head predicts the corresponding scale parameters
$\sigma_\theta$. This yields a stochastic policy with closed-form
log-likelihoods, enabling sampling-based control while retaining the benefits of
the OpenVLA-OFT backbone.

\section{Implementation Details}
\label{sec:implementation}

\subsection{Architecture Hyperparameters}

We summarize the instantiated components of ProGAL-VLA. During inference, the total parameter count is dominated by the OpenVLA backbone; the prospective planner $\pi_{slow}$ operates asynchronously and does not affect control-time latency.

\begin{table*}[t]
\centering
\caption{Hyperparameters for ProGAL-VLA Components.}
\begin{tabular}{l|l}
\toprule
\textbf{Component} & \textbf{Specification} \\
\midrule
\multicolumn{2}{c}{\textit{Prospective Planner} ($\pi_{slow}$)} \\
\midrule
Model Backbone & Qwen-2.5-VL-Instruct-7B \\
Input Resolution & $448 \times 448$ \\
Output Format & Symbolic template (e.g., \texttt{grasp\_green\_mug}) \\
Invocation Frequency & Once per episode (or when grounding fails) \\
\midrule
\multicolumn{2}{c}{\textit{Grounded State Module (GSM)}} \\
\midrule
Object Detector & YOLO-World (L-sized pretrained model) \\
Depth Estimator & Metric3D (v2, ViT-L backbone) \\
Graph Memory Size & $N_{\max}=16$ entity nodes (FIFO update) \\
Entity Embedding Dim. & $d=1024$ \\
\midrule
\multicolumn{2}{c}{\textit{Action Policy} ($\pi_{fast}$)} \\
\midrule
Policy Backbone & OpenVLA-7B (Llama-2 based) \\
Semantic Input & Verified goal embedding $g_t$ projected to 4096-d \\
Action Space & 7-DoF end-effector motion + binary gripper command \\
\midrule
\multicolumn{2}{c}{\textit{Training Configuration}} \\
\midrule
Optimizer & AdamW \\
Learning Rate & $2\times10^{-5}$ with linear warmup \\
Batch Size & 128 \\
GAC Loss Weight $\lambda$ & 0.1 \\
Temperature $\tau$ & 0.07 \\
GPUs & 4 $\times$ NVIDIA A6000 \\
\bottomrule
\end{tabular}
\end{table*}

\section{Extended Experimental Results}
\label{sec:results}

We provide detailed breakdowns that complement the aggregate results in the main text and expose specific robustness properties and failure behaviors of ProGAL-VLA.

\subsection{Granular Robustness Analysis}

Table~\ref{tab:robustness_breakdown} decomposes failures according to the underlying perturbation type. This isolates whether degradation is caused by robot calibration errors, sensor noise, or geometric changes in viewpoint and field-of-view.

\textbf{Robot Perturbations: Calibration Drift vs. High-Frequency Noise.}

\begin{itemize}
    \item \textbf{Joint Drift (systematic offset).}
    OpenVLA collapses to 2.1\% success under drift, indicating heavy reliance on precise proprioceptive–visual alignment. ProGAL-VLA retains 68.2\% success, demonstrating that the verified goal embedding $g_t$ serves as a stable 3D anchor that is independent of small calibration errors.
    
    \item \textbf{Joint Jitter (sensor noise).}
    ProGAL-VLA reaches 74.8\% success, substantially outperforming both baselines. The GSM’s temporal 3D graph filters out high-frequency noise that destabilizes conventional frame-conditioned policies.
\end{itemize}

\textbf{Camera Perturbations: Geometric Invariance.}

\begin{itemize}
    \item \textbf{Viewpoint Shift.}
    OpenVLA degrades to 1.2\% under moderate viewpoint changes, confirming a reliance on 2D pixel coordinates. ProGAL-VLA maintains 85.1\% success, showing robustness to geometric camera motion due to reasoning in 3D entity space.

    \item \textbf{Field-of-View Variation.}
    ProGAL-VLA achieves 91.7\% robustness under FOV changes (zoom in/out), indicating scale invariance in grounding and action generation.
\end{itemize}

\begin{table*}[t]
\centering
\caption{Robustness Breakdown (Success Rate \%). ``Drift'' refers to systematic joint offsets; ``Jitter'' refers to high-frequency noise.}
\label{tab:robustness_breakdown}
\begin{tabular}{l|cc|cc}
\toprule
& \multicolumn{2}{c|}{\textbf{Robot Perturbation}} & \multicolumn{2}{c}{\textbf{Camera Perturbation}} \\
\textbf{Model} & \textit{Joint Drift} & \textit{Joint Jitter} & \textit{Viewpoint Shift} & \textit{FOV Change} \\
\midrule
OpenVLA~\cite{kim2024openvla} & 2.1 & 4.9 & 1.2 & 0.4 \\
OpenVLA-OFT+~\cite{kim2025fine} & 28.5 & 35.1 & 68.4 & 78.0 \\
\textbf{ProGAL-VLA (Ours)} & \textbf{68.2} & \textbf{74.8} & \textbf{85.1} & \textbf{91.7} \\
\bottomrule
\end{tabular}
\end{table*}

\subsection{Failure Mode Analysis}

Table~\ref{tab:failure_modes} dissects failure modes to quantify the “language-ignorance’’ issue and show how the Verification Bottleneck eliminates it.

\textbf{Grounding Errors (Language Ignorance).}

A grounding failure occurs when the agent interacts with a geometrically plausible but semantically incorrect object.

\begin{itemize}
    \item OpenVLA exhibits 41.2\% grounding failures, demonstrating strong bias toward visual shortcuts rather than instruction semantics.
    \item ProGAL-VLA reduces this failure type to 6.3\%, confirming that symbolic-to-entity verification significantly improves semantic fidelity.
\end{itemize}

\textbf{Ablation: Role of Contrastive Alignment.}

The model without the GAC loss,
\texttt{ProGAL (w/o $\mathcal{L}_{GAC}$)}, 
shows that architecture alone does not guarantee correct grounding:
\begin{itemize}
    \item Removing $\mathcal{L}_{GAC}$ increases grounding failures from 6.3\% to 22.4\%.
    \item This demonstrates that maximizing $I(S;E)$ is essential for binding symbolic sub-goals to the correct 3D entities.
\end{itemize}

\textbf{Shift in Error Distribution.}

The baseline is dominated by cognitive grounding failures, whereas ProGAL-VLA’s remaining errors are primarily due to physical execution (grasping). This indicates that the Verification Bottleneck effectively resolves the core semantic alignment problem.

\begin{table*}[t]
\centering
\caption{Failure Mode Distribution (Lower is better). Values are percentages of total episodes.}
\label{tab:failure_modes}
\begin{tabular}{l|c|c|c|c}
\toprule
\textbf{Model} & \textbf{Grounding Fail} & \textbf{Grasp Fail} & \textbf{Planning Fail} & \textbf{Success Rate} \\
\midrule
OpenVLA~\cite{kim2024openvla} & 41.2\% & 12.5\% & 5.1\% & 41.2\% \\
ProGAL (w/o $\mathcal{L}_{GAC}$) & 22.4\% & 14.1\% & 4.8\% & 58.7\% \\
\textbf{ProGAL-VLA} & \textbf{6.3\%} & \textbf{7.1\%} & \textbf{4.5\%} & \textbf{82.1\%} \\
\bottomrule
\end{tabular}
\end{table*}

\section{Formal Specification of the Verification Bottleneck}
\label{sec:verification_formal}

For completeness, we formalize the architectural constraint referred to as the
Verification Bottleneck in the main paper. At each timestep $t$, let the prospective planner $\pi_{slow}$ output a symbolic
sub-goal $s_t$ from the language-vision model. The Grounded State Module (GSM)
maps the observation $O_t$ into a set of perceptual entities
$\{e_{t}^{(1)}, \dots, e_{t}^{(K_t)}\}$ with associated 3D attributes.

The verification function
\[
\phi(s_t, \{e_{t}^{(i)}\})
\rightarrow g_t \in \{e_{t}^{(1)}, \dots, e_{t}^{(K_t)}\}
\]
selects a single grounded entity consistent with the symbolic sub-goal.
The action policy $\pi_{fast}$ receives only the verified entity $g_t$
and does not have access to the instruction $L$ or the symbolic template $s_t$.

This induces the conditional-independence constraint, 
\(
a_t \perp L \mid (g_t, O_t, q_t),
\) which is the Verification Bottleneck.

This section clarifies that the bottleneck is architectural, not statistical:
$\pi_{fast}$ cannot bypass the grounding step, ensuring that semantic alignment
is enforced structurally rather than implicitly.

\section{Entity Memory Update Mechanism in GSM}
\label{sec:gsm_update}

The Grounded State Module maintains a bounded entity memory
$M_t = \{e_{t}^{(1)}, \dots, e_{t}^{(K_t)}\}$ with capacity $N_{\max}=16$.

Given a new observation $O_t$, YOLO-World provides 2D detections
and Metric3D provides depth estimates. Each detection is converted into a 3D
entity embedding with appearance, geometry, and positional attributes.

The update procedure is:

\begin{enumerate}
    \item Extract candidate entities from the current frame.
    \item If $|M_t| < N_{\max}$, append all entities directly.
    \item If memory is full, remove the oldest entries (FIFO) and insert new ones.
    \item The resulting memory $M_t$ forms the node set of the temporal 3D entity graph.
\end{enumerate}

This memory is not used for long-horizon temporal reasoning; instead,
it provides short-range stability and allows $\pi_{fast}$ to operate on a
temporally smoothed representation that suppresses frame-level noise.

\section{Verified Goal Conditioning in $\pi_{fast}$}
\label{sec:goal_conditioning}

The action policy $\pi_{fast}$ conditions exclusively on the verified goal
$g_t$ produced by the GSM and does not receive direct language input. Let $h(g_t)$ be the learned projection of the grounded entity embedding into a
4096-dimensional vector. The policy input at timestep $t$ is,
\(
x_t = h(g_t),
\)
and the control distribution is,
\(
a_t \sim \pi_{fast}(x_t).
\) Because $x_t$ is derived solely from $g_t$, any semantic interpretation of the
language instruction must flow through the verification step. This section
clarifies how the architectural routing enforces the independence relation
used in Proposition 1.

\section{Symbolic Template Resolution}
\label{sec:template_resolution}

The prospective planner $\pi_{slow}$ produces symbolic templates such as
\texttt{grasp\_green\_mug}. These templates are not executed directly; instead,
they index the grounding step.

Given a template $s_t$, a corresponding attribute filter is applied to the
entities in memory:
\[
\Gamma(s_t) = \{ e \in M_t : e \text{ matches attributes in } s_t \}.
\]
If multiple entities satisfy the template, a simple nearest-in-view heuristic
(selecting the entity with largest detection confidence) is used.

The verified goal is, 
\(
g_t = \arg\max_{e \in \Gamma(s_t)} \text{conf}(e).
\) This template-resolution stage provides the symbolic-to-perceptual bridge
that links $\pi_{slow}$ and $\pi_{fast}$ without exposing language features to
the control policy.

\section{Detailed results of LIBERO-Plus}

This section provides a detailed characterization of generalization under distribution shifts in the LIBERO-Plus benchmark. Figure~\ref{fig:libero_robustness} reports success rates under seven perturbation dimensions—Camera, Robot Initialization, Language Instruction, Lighting, Background, Sensor Noise, and Scene Layout—for a wide range of VLA policies, with each bar further decomposed by task suite (Spatial, Object, Goal, and Long-horizon). This factorized view exposes distinct robustness profiles: some methods are relatively stable to robot and layout shifts but degrade sharply under camera or language perturbations, while others show the opposite trend. Our ProGAL-VLA model achieves consistently strong performance across nearly all perturbation types and suites, indicating not only higher average success rates but also more uniform behavior under diverse sources of ambiguity and visual variation.

\begin{figure*}[t!]
    \centering
    % Row 1
    \caption{\textbf{Fine-grained Robustness Analysis on LIBERO-Plus.} 
    We visualize the success rates across seven perturbation dimensions (x-axis) for four task suites (Spatial, Object, Goal, Long). While baselines such as OpenVLA and $\pi_0$ exhibit significant performance degradation under \textit{Camera} and \textit{Robot} perturbations, \textbf{ProGAL-VLA} maintains high robustness across most dimensions.}
    \label{fig:libero_robustness}
    \vspace{-10pt}

    \includegraphics[width=\linewidth]{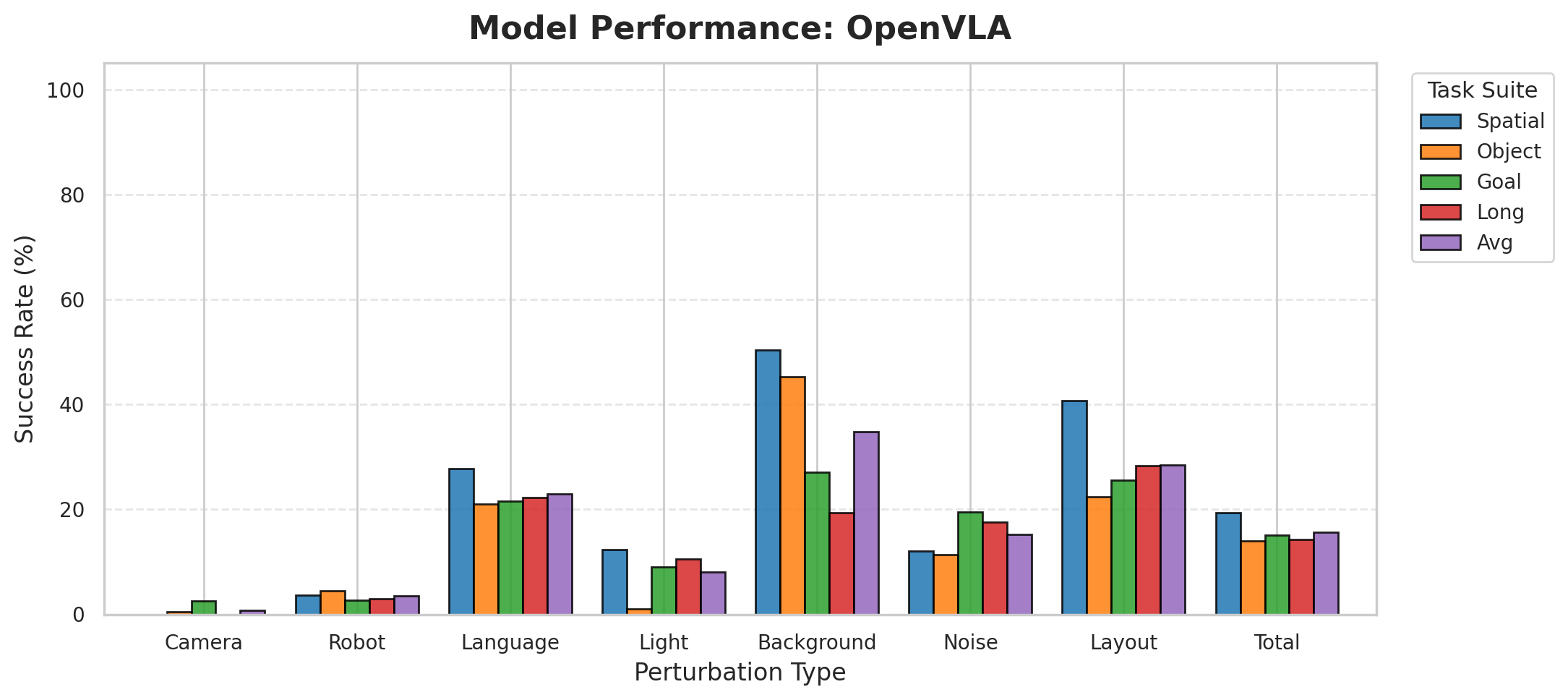}
    \includegraphics[width=\linewidth]{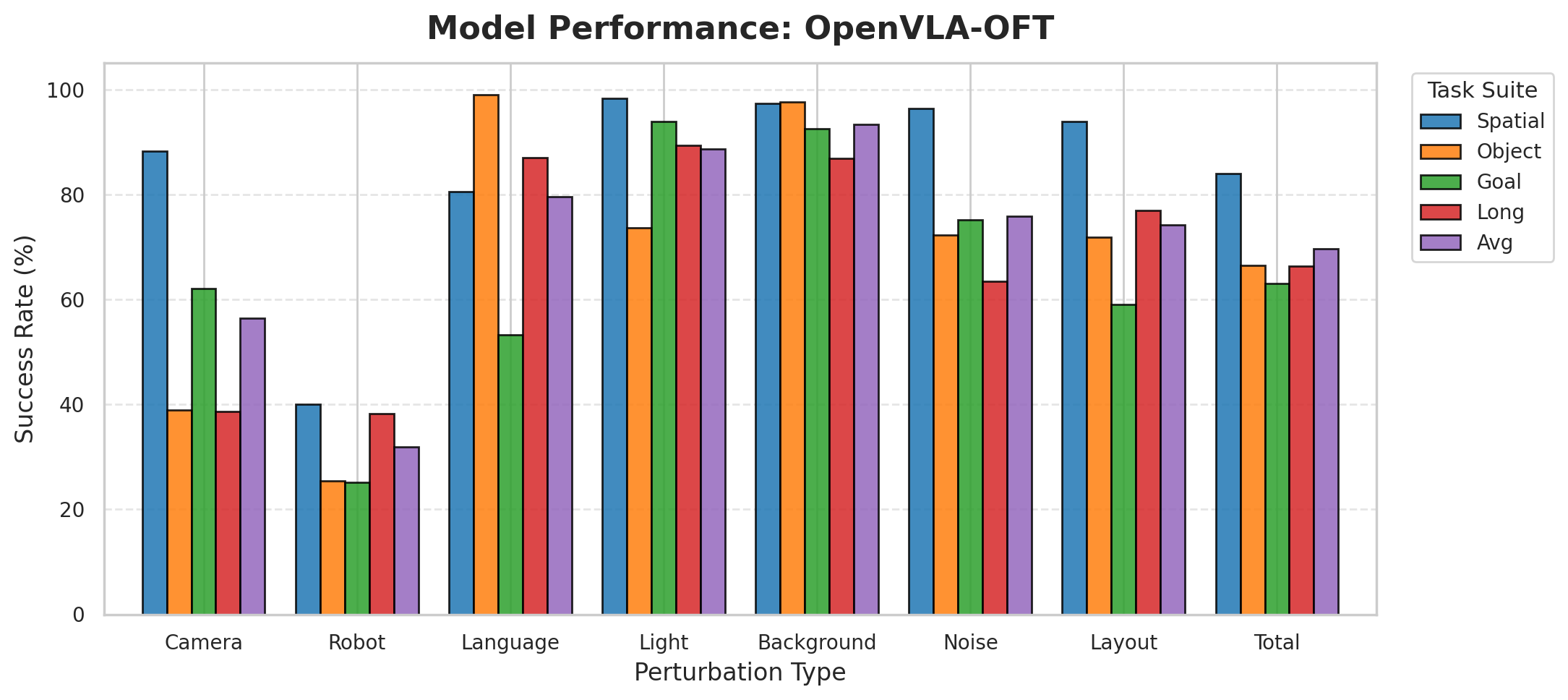}
    \includegraphics[width=\linewidth]{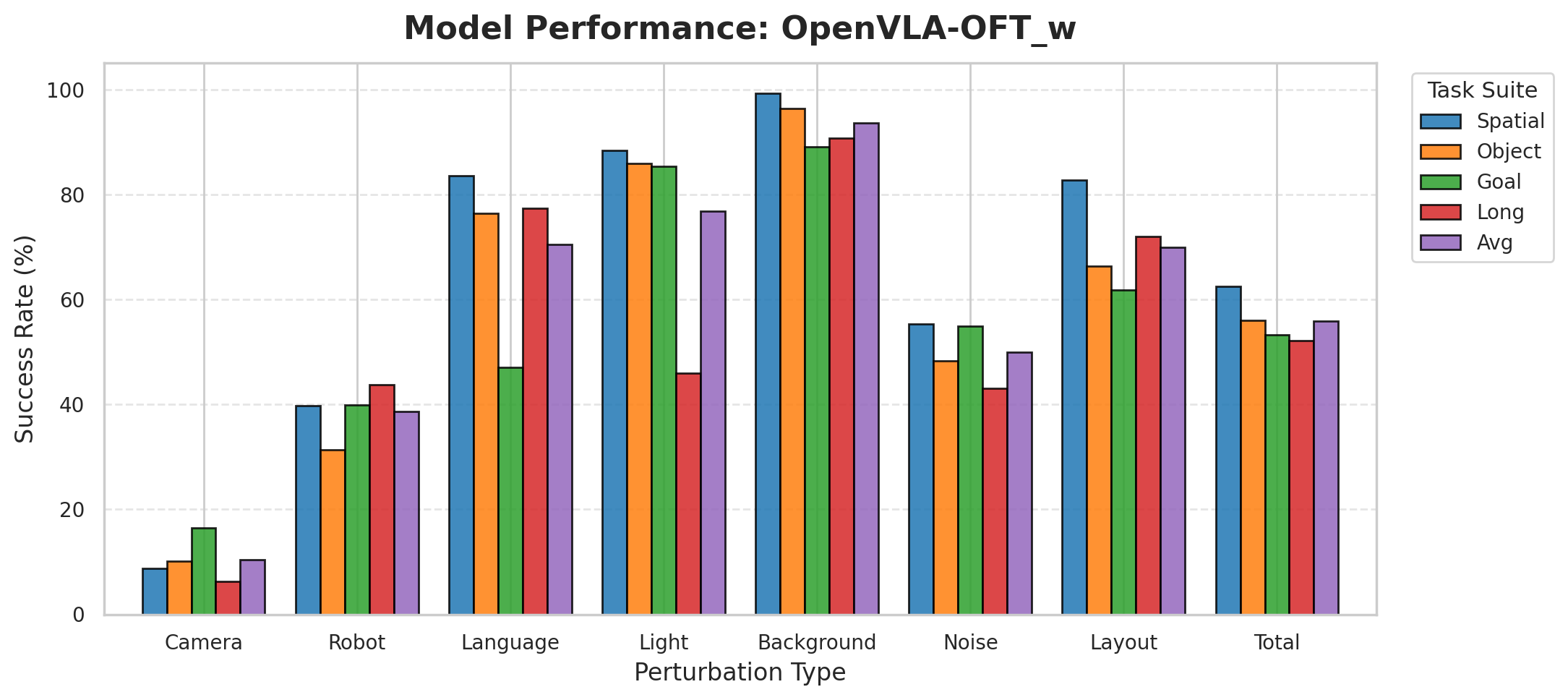}
\end{figure*}

\begin{figure*}[t!]
    \centering
    \ContinuedFloat
    \includegraphics[width=\linewidth]{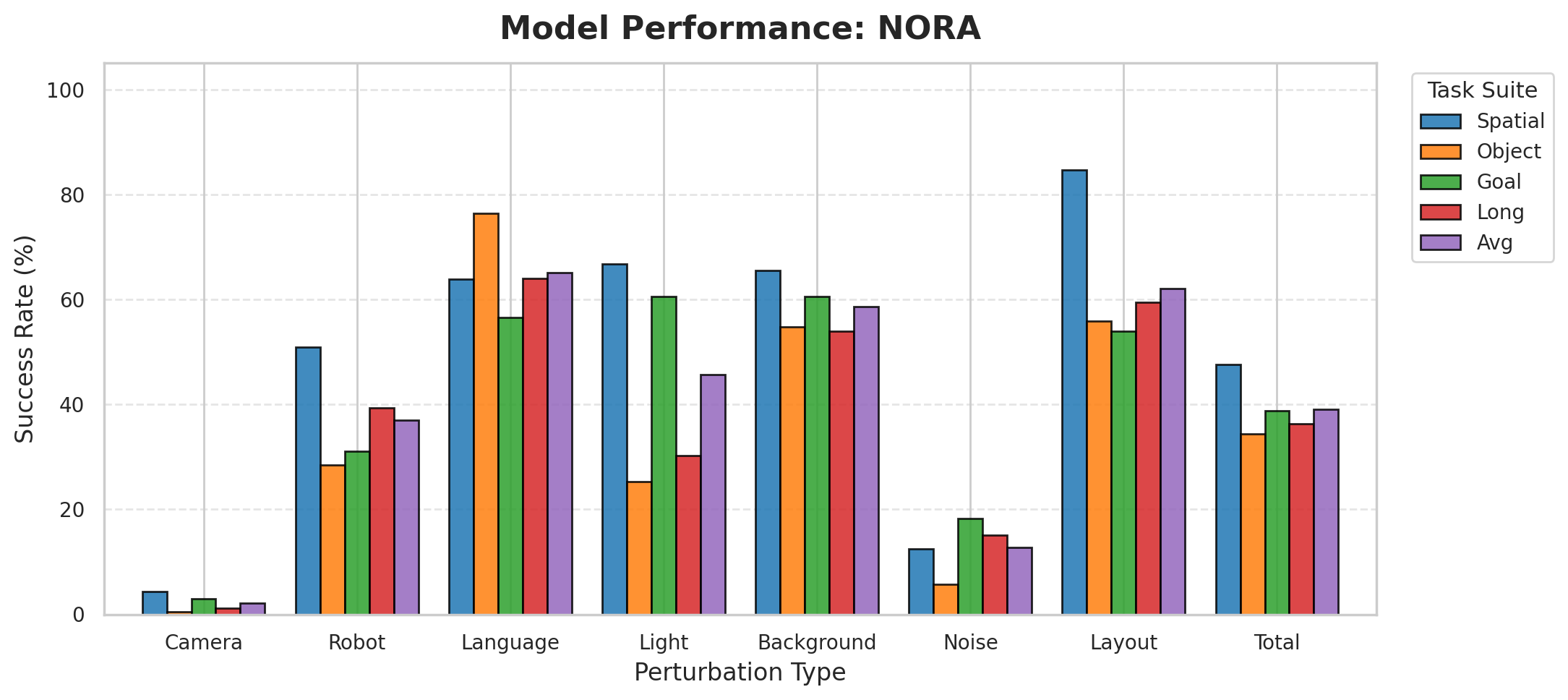}
    \includegraphics[width=\linewidth]{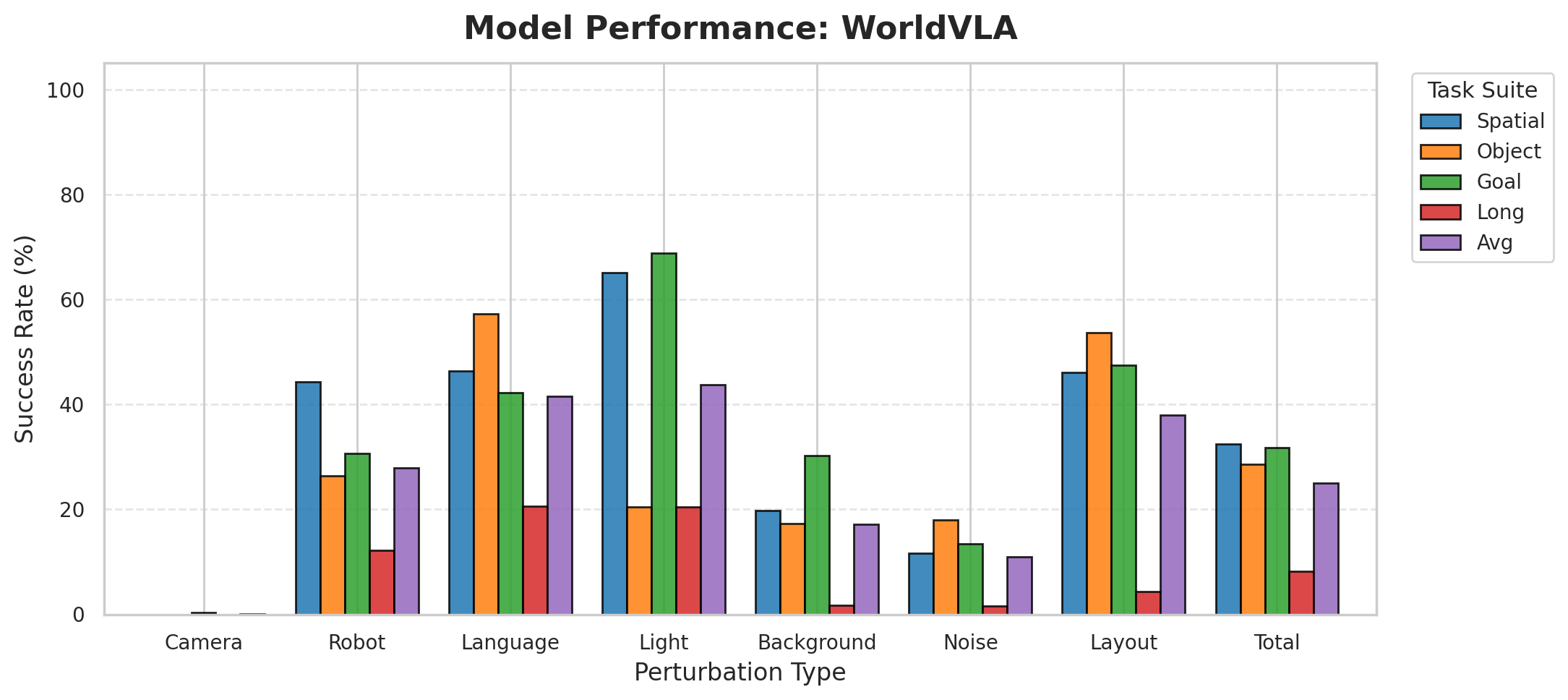}
    \includegraphics[width=\linewidth]{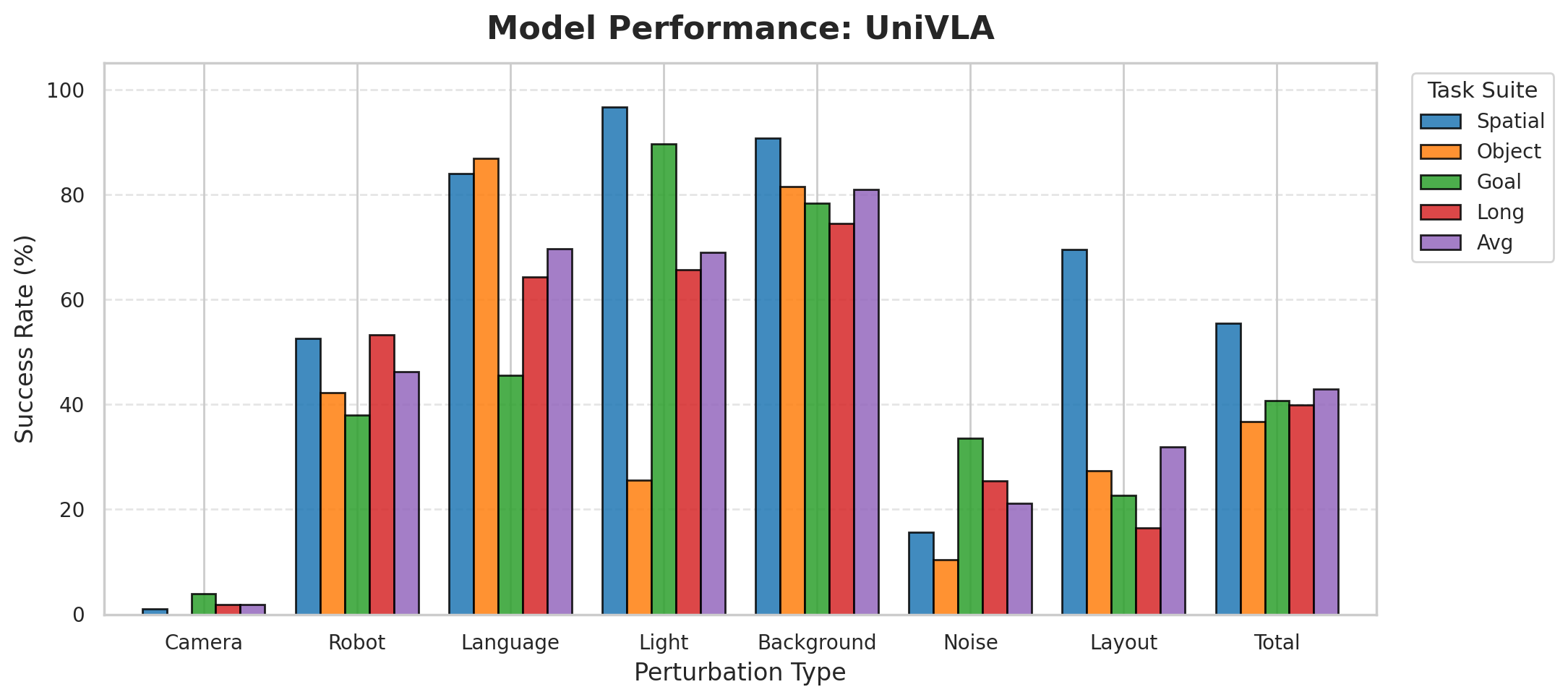}
\end{figure*}

\begin{figure*}[h!]
    \centering
    % Row 1
    \ContinuedFloat
    \includegraphics[width=\linewidth]{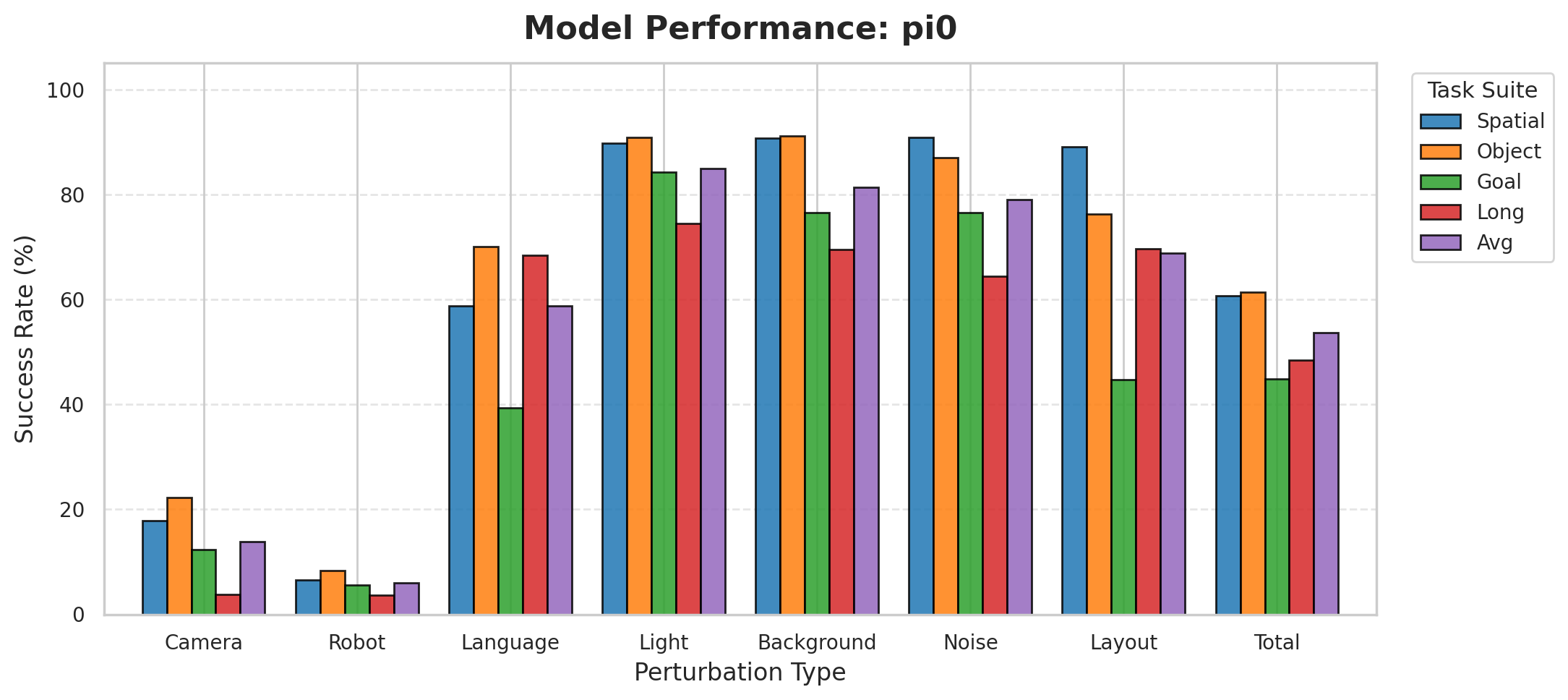}
    \includegraphics[width=\linewidth]{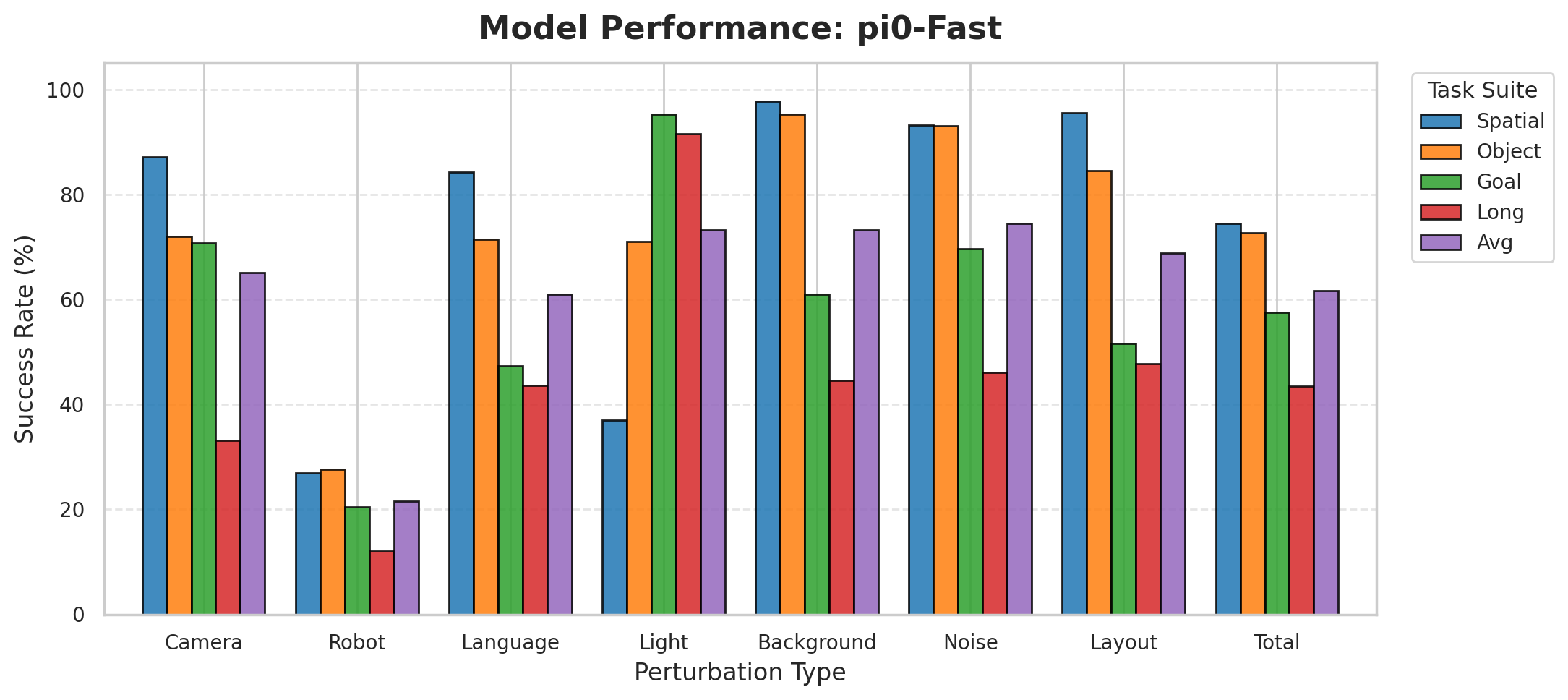}
    \includegraphics[width=\linewidth]{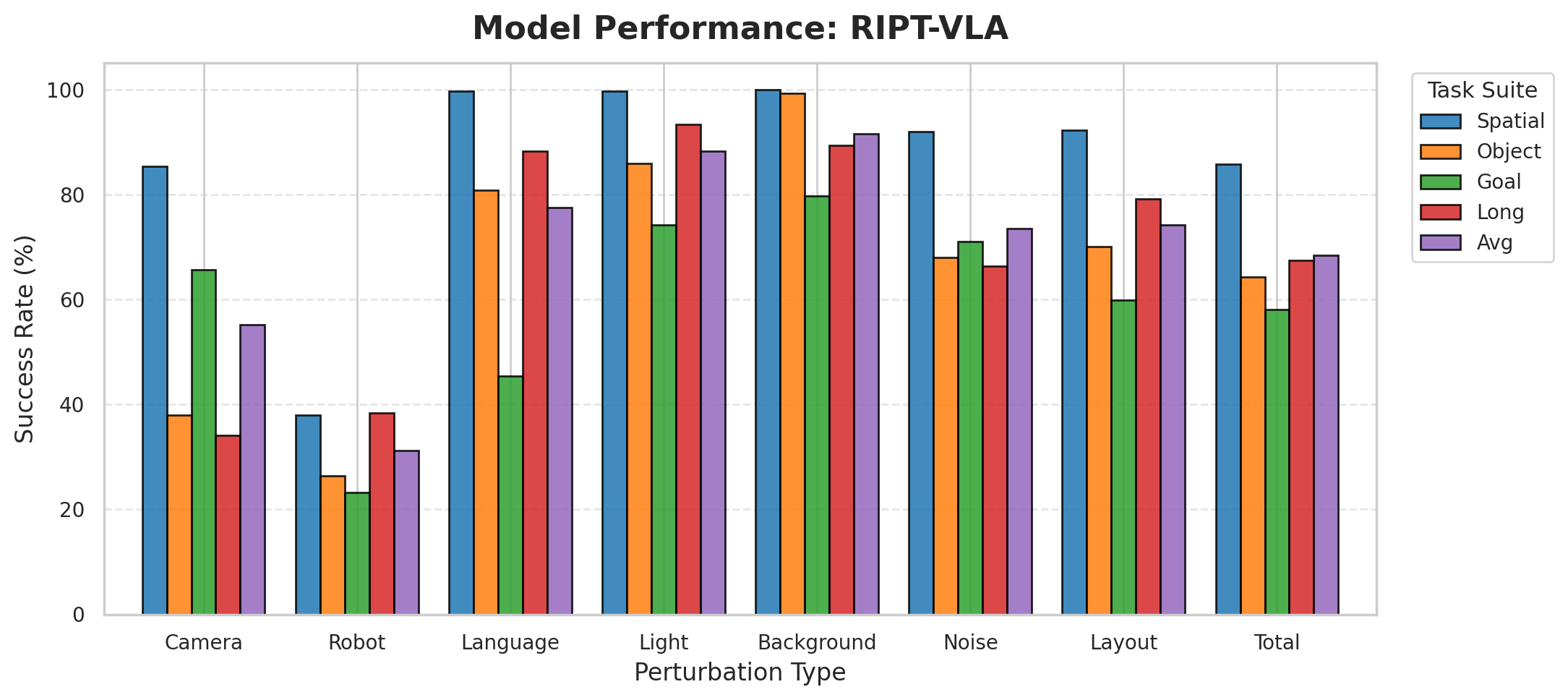}
\end{figure*}

\begin{figure*}[h!]
    \centering
    % Row 1
    \ContinuedFloat
    \includegraphics[width=\linewidth]{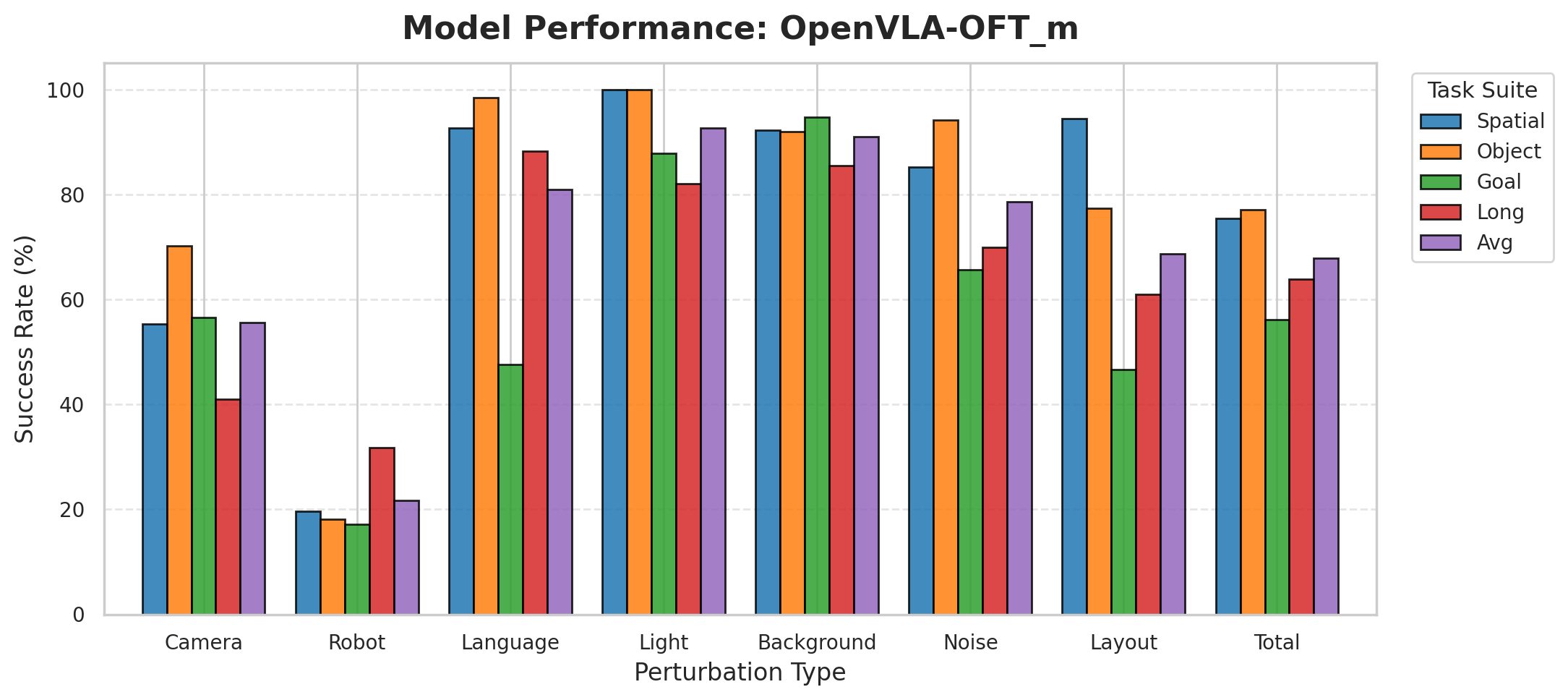}
    \includegraphics[width=\linewidth]{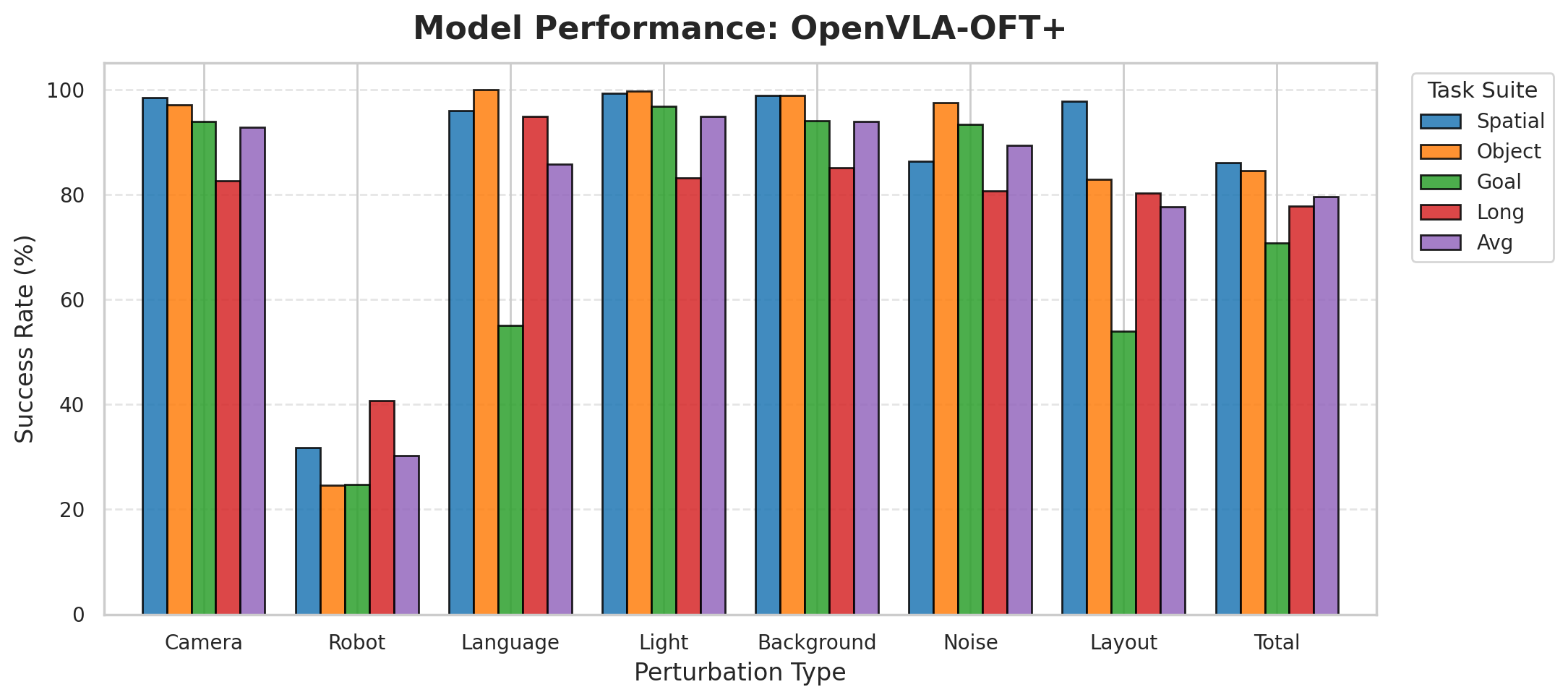}
    \includegraphics[width=\linewidth]{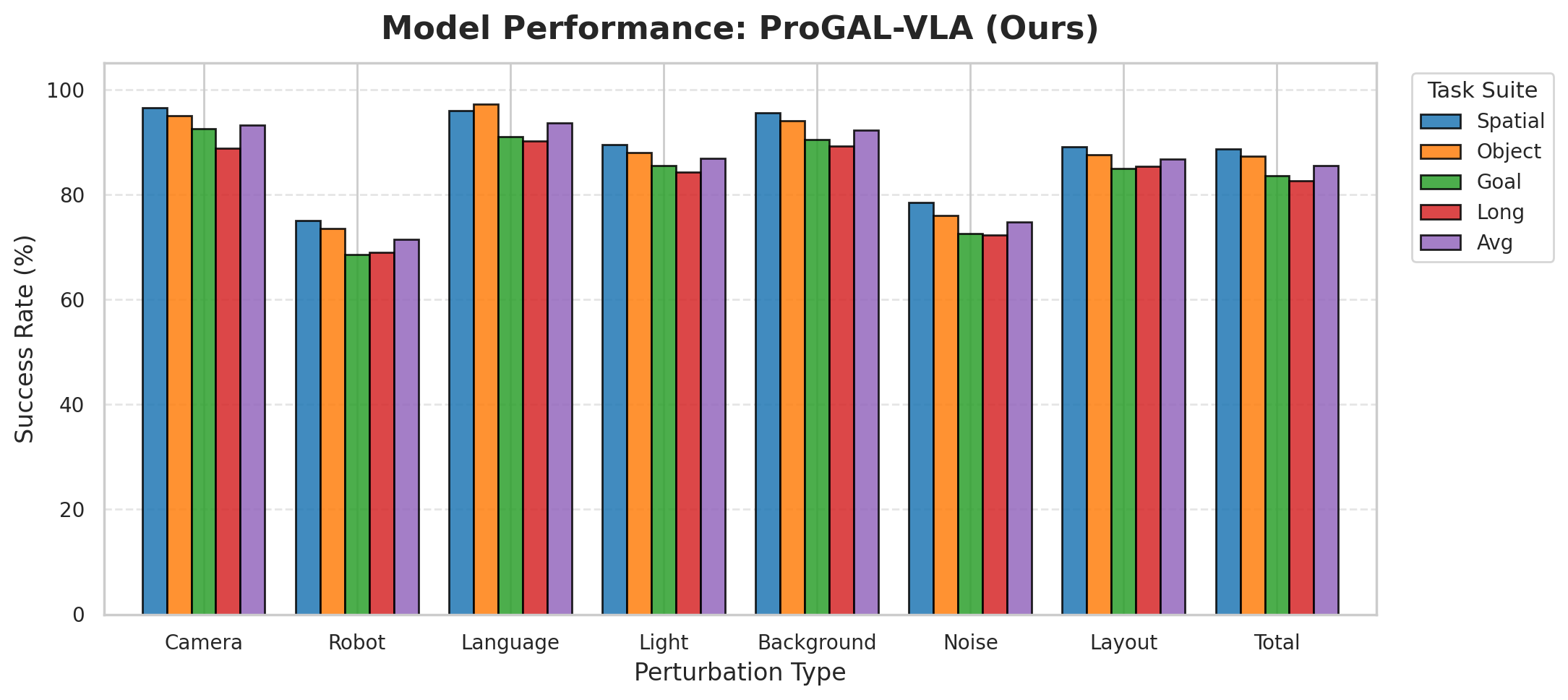}
\end{figure*}

\end{document}